\documentclass[a4paper,11pt]{iopart}

\usepackage{microtype}
\usepackage{graphicx}
\usepackage{subfigure}
\usepackage{booktabs} 

\usepackage{mathtools}
\usepackage{amssymb}
\usepackage{hyperref}

\usepackage{mathtools}
\usepackage{amsthm}
\usepackage{natbib}
\usepackage{xcolor}

\usepackage{setspace}

\usepackage{fontawesome}
\newcommand{\github}{\href{https://github.com/NiallJeffrey/EvidenceNetworksDemo}{\faGithub}}

\mathchardef\mhyphen="2D 
\begin{document} 

\title[]{\large Evidence Networks: simple losses for fast, amortized, neural Bayesian model comparison}


\vspace{-2pt}
\author{Niall Jeffrey$^{1}$ \& Benjamin D. Wandelt$^{2,3}$}
\address{\scriptsize $^{1}$ Department of Physics \& Astronomy, University College London, Gower St., London \\ $^{2}$ Institut d'Astrophysique de Paris (IAP), UMR 7095, CNRS, Sorbonne Universit\'e, Paris\\
$^{3}$ Center for Computational Astrophysics, Flatiron Institute, 162 5${\rm th}$ Avenue, New York}
\ead{n.jeffrey@ucl.ac.uk}
\begin{indented}
\vspace{2pt}
\item[]Original: April 2023. Revised: August, 2023.
\end{indented}
\vspace{-2pt}

\begin{abstract}
Evidence Networks can enable Bayesian model comparison when state-of-the-art methods (e.g. nested sampling) fail and even when likelihoods or priors are intractable or unknown. Bayesian model comparison, i.e. the computation of Bayes factors or evidence ratios, can be cast as an optimization problem. Though the Bayesian interpretation of optimal classification is well-known, here we change perspective and  present classes of loss functions  that result in fast, amortized neural estimators that directly estimate convenient functions of the Bayes factor. This mitigates numerical inaccuracies associated with estimating individual model probabilities.  We introduce the leaky parity-odd power (l-POP) transform, leading to the novel ``l-POP-Exponential'' loss function.  We explore neural density estimation for data probability in different models, showing it to be less accurate and scalable than Evidence Networks. Multiple real-world and synthetic examples illustrate that Evidence Networks are explicitly independent of dimensionality of the parameter space and scale mildly with the complexity of the posterior probability density function. This simple yet powerful approach has broad implications for model inference tasks. As an application of Evidence Networks to real-world data we compute the Bayes factor for two models with gravitational lensing data of the Dark Energy Survey. We briefly discuss applications of our methods to other, related problems of model comparison and evaluation in implicit inference settings. 
\end{abstract}
\vspace{-0.25cm}

%


\vspace{-0.1cm}
\section{Introduction}
\label{Intro}  \vspace{-0.2cm}
\enlargethispage*{0.5cm}
\subsection{Background} \vspace{-0.15cm} Bayesian model comparison aims to evaluate the relative posterior probability of different underlying models given some observed data $x_O$. The ratio of probabilities (the \textit{posterior odds}) for two models, $M_1$ and $M_0$, can be related to the \textit{model evidence} via:
\begin{equation}
\frac{p(M_1 | x_O)}{p(M_0 | x_O)} = \frac{p(x_O | M_1 )}{p( x_O | M_0 )} \ \frac{p( M_1 )}{p( M_0 )} \ \ ,
\end{equation}
\noindent where $p(x_O | M_1 )$ is the \textit{evidence} for model $M_1$ and $p(M_1)$ is the prior probability of the model. The evidence ratio for the two models is known as the Bayes factor, $K$ (see~\citealt{jaynes07} for details). 

Under the assumption of equal prior probability for both models, the Bayes factor (i.e. the ratio of model evidences) becomes equal to the ratio of posterior probabilities for models:
\begin{equation}
 K = \frac{p(x_O | M_1 )}{p( x_O | M_0 )} = \frac{p(M_1 | x_O)}{p(M_0 | x_O)} \ ,  \ \ {\rm if \ } p( M_1 )=p( M_0 ) \ \ .
\end{equation}

The evaluation of the Bayes factor, given some data and taking into account all sources of uncertainty, is typically the ultimate goal for model comparison. Bayes factors have been used or discussed in the fields of epidemiology, engineering, particle physics, cosmology, law, neuroscience, and many others~\citep{wakefield2009bayes, yuen2010recent, acoustics, 6753897, bayes_law, Handley_2019, keysers2020using, massimi}. Though the simplest interpretation is that of an odds ratio under equal model priors, interpretations like the \textit{Jeffreys' scale}~\citep{jeffreys1998theory} are also popular for Bayesian model comparison. In general these interpretations are concerned with the log Bayes factor, thought of as the relative magnitude of the odds.

\subsection{Paper summary}
In section~\ref{sec:motivation}, `{\hypersetup{hidelinks}\nameref{sec:motivation}}', we introduce the challenges faced in Bayesian model comparison that are solved by Evidence Networks: intractability of computationally challenging integration, unknown likelihood/prior, and lack of model parameterization. In section~\ref{sec:compare}, `{\hypersetup{hidelinks}\nameref{sec:compare}}', we consider existing methods in the context of these three motivations.

In section~\ref{Optimization}, `{\hypersetup{hidelinks}\nameref{Optimization}}', we show how to construct bespoke losses to estimate convenient functions of the model evidence, with focus on the log Bayes factor. Section~\ref{sec:direct_posterior_loss} describes the first case of the `{\hypersetup{hidelinks}\nameref{sec:direct_posterior_loss}}'. Section~\ref{sec:direct_bayes_loss} introduces our primary case of `{\hypersetup{hidelinks}\nameref{sec:direct_bayes_loss}}', which includes the description of the novel ``l-POP'' transform and associated ``l-POP-Exponential'' loss. We include a `{\hypersetup{hidelinks}\nameref{sec:summary_symmetric}}' in section~\ref{sec:summary_symmetric}.

In section~\ref{sec:pitfalls}, we discuss `{\hypersetup{hidelinks}\nameref{sec:pitfalls}}' with Evidence Networks. Section~\ref{sec:pitfall_challenge} describes the practical `{\hypersetup{hidelinks}\nameref{sec:pitfall_challenge}}' in validating a trained Evidence Network in the standard case in which the true Bayes factor of the validation data is not known. Section~\ref{sec:pitfall_solutions} presents a `{\hypersetup{hidelinks}\nameref{sec:pitfall_solutions}}' solution.

section~\ref{sec:timeseries} presents `{\hypersetup{hidelinks}\nameref{sec:timeseries}}'. This demonstration uses a `{\hypersetup{hidelinks}\nameref{sec:genmodel}}' (section~\ref{sec:genmodel}) defined such that the Bayes factors can be calculated analytically using a closed-form expression. Section \ref{sec:timeseries_results}, `{\hypersetup{hidelinks}\nameref{sec:timeseries_results}}', describes the results of the Evidence Networks, demonstrates the blind coverage test, and explores the choice of loss functions. Alternatives are considered in section \ref{sec:polychord}, `{\hypersetup{hidelinks}\nameref{sec:polychord}}', and section \ref{sec:density_method}, `{\hypersetup{hidelinks}\nameref{sec:density_method}}'.

section~\ref{sec:des} presents `{\hypersetup{hidelinks}\nameref{sec:des}}', in which Evidence Networks are used to compare two alternative models of galaxy intrinsic alignments using gravitational lensing data.

section~\ref{sec:rastrigin}, `{\hypersetup{hidelinks}\nameref{sec:rastrigin}}', addresses a conceptual point: there are cases where estimation of the Bayes factor (e.g. with Evidence Networks) for model
comparison can be done even in cases where solving the parameter inference problem is intractable, e.g. due to the complexity of the posterior probability distribution.

We conclude in section~\ref{sec:conclusion}, with section~\ref{sec:extensions} presenting a discussion of `{\hypersetup{hidelinks}\nameref{sec:extensions}}': absolute evidence (rather than Bayes factor ratios), the connection to frequentist hypothesis testing, and Evidence Networks for posterior predictive tests.

\subsection{Motivation}\label{sec:motivation}The challenge for Bayesian model comparison arises in the calculation of the model evidence. Assuming the statistical model is known, the model evidence can be expressed as the \textit{marginal likelihood}:
\begin{equation} \label{eq:marg_ev}
p(x_O | M_1) = \int p(x_O | \theta, M_1 ) \ p(\theta | M_1 ) \  \mathrm{d} \theta \ \ .
\end{equation}
\noindent This integral over the space of permissible model parameters $\theta$ is often intractable. There could be three distinct reasons for this:
\begin{enumerate}
    \item The integral is computationally challenging. This often renders the calculation effectively impossible for a high-dimensional parameter space. 
    \item The probability densities $p(x_O | \theta)$ and $p( \theta)$ are unknown.
    \item There is no known underlying parameterization that generates the data. For example, real-world data can be drawn from different model classes, and there may be no evident underlying parameterization.
\end{enumerate}

\noindent We expand upon these three reasons in section~\ref{sec:compare} below. However, for practitioners used to the marginal likelihood presentation of Bayesian evidence (equation~\eqref{eq:marg_ev}), the key points to  consider first are: \begin{itemize}
    \item  \textit{model evidence does not require an integration over model parameters.} 
    \item \textit{model evidence does not require knowledge of the unnormalized posterior distribution}. 
\end{itemize}
\noindent The marginal likelihood is a technique to marginalise the joint distribution $p(x_O, \theta | M_1)$ to recover $p(x_O | M_1)$, however a knowledge of $p(x_O, \theta | M_1)$ is not actually necessary to evaluate the model evidence $p(x_O | M_1)$. Evidence Networks provide an alternative approach that avoids equation~\eqref{eq:marg_ev} entirely.

\subsection{Evidence Networks \& comparison with existing methods} \label{sec:compare}

Evidence Networks are fast amortized neural estimators of Bayes factors for model comparison. In section~\ref{Optimization} we describe classes of loss functions that lead to the correct optimization objective for achieving this goal. In section~\ref{sec:pitfalls} we describe the potential pitfalls and validation techniques that are applicable to real-world applications.

We will now address the three motivations highlighted above, comparing to existing methods and linking to our demonstrations that are presented within this paper. 
\paragraph{Motivation (i) - intractable integral:} Assuming the integrand of equation~\eqref{eq:marg_ev} is known, the marginal likelihood integral can be intractable in practice. This is despite the large number of computational techniques have been devised to address this problem:  straightforward estimation of the evidence integral, Reversible Jump Markov Chain Monte Carlo, the  Laplace or Saddle Point Approximation, (annealed) Importance Sampling, Path Sampling, Parallel Tempering, Thermodynamic Integration,  and Nested Sampling (for reviews see \citealt{doi:10.1080/01621459.1995.10476572,doi:10.1198/016214501753208780,2014arXiv1411.3013K}).

Consider nested sampling, as it is one of the most popular methods for evidence calculation~\citep{multinest,polychord}; even this approach becomes onerous for high-dimensional parameter spaces, when $\mathrm{dim}(\theta) \gtrsim 10^2$. We will show that the performance of Evidence Networks does not suffer when there are large numbers of parameters as the evaluation of this integral is side-stepped.

\paragraph{Motivation (ii) - unknown likelihood/prior:}  Beyond the challenge of evaluating equation~\eqref{eq:marg_ev}, the likelihood $p(x_O | \theta, M_1 )$ or prior $p(\theta | M_1 )$ is not available for many inference tasks. In the implicit inference framework, which includes likelihood-free or simulation-based inference, these unknown distributions can be learned from training data~\citep{Papamakarios_lfi, delfi1, Brehmer5242, Cranmer201912789}. This parameter inference approach that has enjoyed recent success with scientific applications~\citep{delfi2, Taylor_2019, Brehmer_2019, des_lfi, ramanah2020dynamical,2020arXiv201008537L}. We discuss three alternative options for Bayesian model comparison in this setting.

\begin{itemize}
\item One option is to first estimate the unknown distribution  and then perform the marginal likelihood integration, equation~\eqref{eq:marg_ev}. For example, one might first estimate $p(x | \theta, M_1 )$ from simulated data; this is is known as Neural Likelihood Estimation (NLE). This type of approach has been explored in~\cite{bayes_harmonic}, which leverages a harmonic mean estimator that can be applied, for example, to the learned distributions. However, learning intermediate probability densities in terms of model parameters $\theta$ is an additional and unnecessary step, which scales exponentially poorly with the number of model parameters (the curse of dimensionality). Evidence Networks avoid this poor scaling because they do no ``see'' the model parameters as the parameters do not appear in the expression for the Bayes factor.

Typically it also is necessarily to reduce the data dimensionality through compression to make the density estimation tractable. Since this changes the dimensionality of the evidence probability density function, this can significantly alter the Bayesian model evidence values.  By comparison, with Evidence Networks we can also use uncompressed data, which preserves the correct Bayes factor estimates.

\item Another little-explored option for simulation-based (likelihood-free) inference is direct density estimation of the probability density $p(x | M_1)$, for which the Bayesian factor is just the density ratio evaluated for the observed data: $\frac{p(x_O | M_1)}{p(x_O | M_1)}$.

\indent This approach is appealing as it also avoids the marginal likelihood integration to directly estimate the Bayes factor for model comparison. However, this approach suffers from the curse of dimensionality for high-dimensional data. This is explored in section~\ref{sec:density_method}.

\item Direct estimation of the model evidence (or model posterior) corresponds to the $\mathcal{M}$-closed limit (i.e. all possible models are accounted for, so that $\sum_i p(M_i) = 1$) as described in~\cite{evidential}. In this work we explore alternative forms of the loss. We find that for model comparison, rather than taking ratios of individual model posterior estimates, it is better to estimate the Bayes factors directly using a bespoke loss function.

\end{itemize}

\paragraph{Motivation (iii) - no parameterization:}  For many real-world model comparison tasks the underlying parameterization is itself unknown. A set of natural images, draws from a generative model, or human-generated data have no obvious model parameters\footnote{The weights and biases of neural generative models are parameters, but so numerous that  Bayesian Model comparison via marginal likelihood is intractable.}.   Evidence Networks do not need to refer to model parameters to compute Bayes factors and therefore enable Bayesian model comparison even for machine-learned models. The expression for Bayesian evidence (e.g. $p(x_O| M_1)$) makes no reference to parameters. This observation provides a clue that parameters can be bypassed entirely in evaluating Bayes factors.

\section{Optimization objective}
\label{Optimization}

We will now show how to design classifier training losses that lead to networks that estimate convenient functions of the model evidence (e.g. the log Bayes factor) for a given data set. 

We assume training data drawn from $p(x | \mathcal{M})$ for different models $\mathcal{M}$, e.g. for model $M_1$:
\begin{equation}
    x_i \sim p(x | M_1)
\end{equation}
In the case of simulated data, for the optimization objective to recover the correct model posterior or Bayes factor estimates, we can assume any underlying model parameters $\theta$ are drawn from the desired prior distribution, e.g. for model $M_1$: 
\begin{equation}
\begin{split}
\theta_i &\sim p(\theta | M_1) \\ 
x_i &\sim p(x | \theta_i, M_1) \ \ .
\end{split}
\end{equation}
\noindent These model parameter values are then not used by the Evidence Network, though their distribution is the implicit prior in the final model evidence or Bayes factor estimation. If we do want to use a different parameter prior $p(\theta)$ distribution this can be taken into account with appropriate $\theta$-dependent weights in the loss function (akin to importance sampling). The training set consists only of data paired with model labels.

The fraction of each model label in the training data acts as an implicit prior of the models. For simplicity, we can assume $p( M_1) = p( M_0)$, which corresponds to each model having equal number of samples in the training data. If a model is over-represented in the training data, or one indeed prefers a given model \textit{a priori}, then a simple re-weighting can be applied to the result.

For a given loss function $\mathcal{V}$, the global optimization objective is given by the functional
\begin{equation} \label{eq:objective}
    I[f] = \sum_{m \in \{0,1\}} \int \ \mathcal{V}(f(x), m) \ p(x, m) \  \mathrm{d}x  \ \ ,
\end{equation}
\noindent acting on functions $f$ of the data, using model labels $m \in \{0,1\}$ for a pair of models $M_0$ and $M_1$. Let $f^*$ be the global optimal function, i.e. that which minimizes $I$; this will be approximated with a neural network. Certain specific loss functions $\mathcal{V}$ yield optimal functions $f^*$ that are estimates of specific useful functions of the model evidence. Overall, we recommend the \textit{l-POP-Exponential Loss} (described in section~\ref{sec:pop_loss}) to estimate the log Bayes factor. 

\subsection{Direct estimate of model posterior {probability}} \label{sec:direct_posterior_loss}
Though we favour direct estimation of functions of the Bayes factor, we will first review simple optimization objectives that result in neural network estimates of the posterior model probability $p(M_i|x_O)$. 

The first of these cases, the Squared Loss, can be interpreted as simple special case of parameter inference (e.g. in the context of Moment Networks:~\citealt{momentnets}). The alternative case, the Cross-Entropy Loss, matches the $\mathcal{M}$-closed limit (i.e. all possible models are accounted for such that $\sum_i p(M_i) = 1$) as described in~\cite{evidential}. In section~\ref{sec:timeseries} we will show that, in comparison to direct Bayes factor estimation, this approach of direct model posterior estimation is not ideal for model comparison.

\paragraph{Squared \& Polynomial Loss:} 

First let us consider the typical Squared Loss
\begin{equation}
    \mathcal{V}(f(x), m) = (f(x) - m )^2 \ \ .
\end{equation}
\noindent  The functional to be optimized from equation~\eqref{eq:objective} is then given by:
\begin{equation}
    I[f] = \int \ (f(x) - 1 )^2 \ p(x, M_1) + f(x)^2 \ p(x, M_0) \  \mathrm{d}x \ \ .
\end{equation}
\noindent If we minimize this objective using standard methods from variational calculus, i.e. $\delta I = 0$, we recover the solution for the global optimal function:
\begin{equation}
    f^*(x) = p(m=1 | \ x) = p(M_1 | \ x) \ \forall \ x \ \ ,
\end{equation}
\noindent under the assumption that there are two models (so that $p(M_1 | x) = 1 -  p(M_0 | x)$). Therefore, the optimal function, evaluated for the observed data $x_O$, gives the model posterior for model $M_1$: 
\begin{equation}
f^*(x_O) = p(M_1 | \ x_O) \ \ .
\end{equation}
\noindent We can further generalize to a Polynomial Loss with a power $\alpha \in \mathbb{R}$, $\alpha\neq1$,
\begin{equation}
    \mathcal{V}(f(x), m) = m(1-f(x))^\alpha + (1-m) f(x)^\alpha \ \ ,
\end{equation} we recover the same result as for the Squared Loss, but with the model posterior probability raised to a power. This results in:
\begin{equation}
    f^*(x_O) =  p(M_1 | \ x_O)^\frac{1}{\alpha-1} \ \ ,
\end{equation}
\noindent where $\alpha=2$ corresponds to the Squared Loss result, as expected. The choice of $\alpha$ leads to different loss landscapes during training, meaning $\alpha$ is a possible hyper-parameter that can be tuned depending on the use-case.

\paragraph{Cross-entropy loss:} If we take the loss function
\begin{equation}
    \mathcal{V}(f(x), m)  =  m \log \big(f(x) \big)   + (1 - m) \log \big( 1 - f(x) \big)  \ \ ,
\end{equation}
\noindent following the same procedure as before gives the result:
\begin{equation}
    f^*(x_O) = p(M_1 | \ x_O) \ \ .
\end{equation}

\subsection{Direct estimate of Bayes factors} \label{sec:direct_bayes_loss}

\paragraph{Exponential Loss:}
Taking the loss function
\begin{equation} \label{eq:exploss}
\begin{split}
    \mathcal{V}(f(x), m)  &=  e^{(\frac{1}{2} - m )f(x)} \\
\end{split}
\end{equation}
gives the global optimization objective:
\begin{equation}
\begin{split}
    I[f] &= \int \ e^{-\frac{1}{2}f(x)}  \ p(x , M_1)  \ +   e^{\frac{1}{2}f(x)} \ p(x , M_0) \  \mathrm{d}x  \ \ .
\end{split}
\end{equation}
\noindent Following the same procedure as previously, i.e. minimizing this objective using variational calculus, recovers the optimal function:
\begin{equation}
    f^*(x) = \log \Big( \frac{p(M_1 | \ x)}{p(M_0 | \ x)} \Big) \ \forall \ x. 
\end{equation}
\noindent Therefore, the optimal function evaluated for our observed data is:
\begin{equation}
   f^*(x_O) =  \log \Big( K \ \frac{p(M_1 )}{p(M_0 )} \Big) = \log K  +  \log \Big( \frac{p(M_1 )}{p(M_0 )} \Big)\ \ .
\end{equation}
\noindent If the model priors are equal (having the same number of examples of each model label in the training data for the neural network), then the second term vanishes.
\paragraph{Logistic loss:}
An alternative form gives the same final result as the Exponential Loss, with
\begin{equation}
    \mathcal{V}(f(x), m)  = \log \big( 1 + e^{(1-2m)f(x)} \big) \ \ .
\end{equation}
\noindent Following the same optimization procedure as before once again results in an estimate of the log Bayes factor with an additional model prior ratio term:
\begin{equation}
   f^*(x_O) = \log K  +  \log \Big( \frac{p(M_1 )}{p(M_0 )} \Big)\ \ .
\end{equation}

\paragraph{$\alpha$-Exponential and log-$\alpha$-Exponent:} Analogously to the generalization from Squared Loss to Polynomial Loss with the $\alpha$ hyper-parameter, we can introduce a power of $\alpha$ to the losses that directly estimate the Bayes factor or log-Bayes factor.

The ``$\alpha$-Exponential'' Loss is given by:
\begin{equation}
    \mathcal{V}(f(x), m)  = \big( 1 + e^{(1 - 2m )f(x)} \big)^\alpha.
\end{equation}
\noindent Following the same optimization procedure as before, we recover results the estimate of the rescaled log Bayes factor with an additional model prior ratio term:
\begin{equation}
\begin{split}
   f^*(x_O) &= \log K^\frac{1}{1+\alpha}  +  \log  \Big( \frac{p(M_1 )}{p(M_0 )}\Big)^{\frac{1}{1+\alpha}} \ \ , \\ 
   &= \frac{1}{1+\alpha} \log K + \frac{1}{1+\alpha} \log \Big( \frac{p(M_1 )}{p(M_0 )} \Big)  \ \ .
\end{split}
\end{equation}

\noindent Similarly, we can construct the ``$\alpha$-log-Exponential'' Loss: 
\begin{equation}
    \mathcal{V}(f(x), m)  = f(x)  ^{(\frac{1}{2}-m)\alpha} .
\end{equation}
\noindent For this case, we analogously recover the Bayes factor to a power, multiplied  by a model prior ratio factor:
\begin{equation}
   f^*(x_O) =   K^\frac{1}{1+\alpha} \times \Big( \frac{p(M_1 )}{p(M_0 )} \Big)^{\frac{1}{1+\alpha}}  \ \ .
\end{equation}

\paragraph{l-POP-Exponential Loss:} \label{sec:pop_loss}

Finally, we present the ``l-POP-Exponential Loss'', which is our recommended choice of default loss for the Evidence Network. This loss applies a bespoke transformation to the network output followed by the usual Exponential Loss. This is completely equivalent to defining a final layer activation function; for this discussion we include the transformation within the loss to aid interpretability. 

We first define the \textit{parity-odd power} transformation, as:
\begin{equation}
    \tilde{\mathcal{J}}_\alpha (x) \coloneqq x \ |x|^{\alpha-1} \ \ \forall \ x \in \mathbb{R}  \ {\rm where} \ \alpha \in \mathbb{R} \ {\rm and} \ \alpha \ge 1 \ \ .
\end{equation}
\noindent Using the sign function ($\rm sgn$), another way of representing the parity-odd power transformation is:
\begin{equation}
    \tilde{\mathcal{J}}_\alpha (x) = {\rm sgn}(x) \ |x|^{\alpha} \ \ .
\end{equation}
\noindent This function is differentiable everywhere. Note, however, that the intermediate ${\rm sgn}$ or absolute value functions are not differentiable at $x=0$, so care should be taken in code implementation.

Though the parity-odd power transform $\tilde{\mathcal{J}}$ works well, its value and gradient remains close to zero, an effect which increases with increasing $\alpha>1$. We therefore use the \textit{leaky parity-odd power}, or l-POP, transform, which we define as:
\begin{equation}
    {\mathcal{J}}_\alpha (x) \coloneqq x + x \ |x|^{\alpha-1} \ \ .
\end{equation}
We could choose to generalize this further, using $\beta x + x \ |x|^{\alpha-1}$, though we effectively choose $\beta=1$ for all that follows. The resulting parity-odd power Exponential Loss, or ``l-POP Exponential Loss' is given by: 
\begin{equation} \label{eq:popexploss}
    \mathcal{V}(f(x), m)  =  e^{(\frac{1}{2} - m )\mathcal{J}_\alpha (f(x))} \ \ .
\end{equation}
\noindent As discussed above, this is just equivalent to using $\mathcal{J}_\alpha$ as a novel choice of activation function in the network output layer. 

Following the same optimization procedure as before, we recover a rescaled estimate of the log Bayes factor plus an additional model prior ratio term:
\begin{equation}
\begin{split}
   f^*(x_O) &=  \mathcal{J}^{-1}_\alpha  \Big( \log K + \log  \frac{p(M_1 )}{p(M_0 )} \Big) \\
   &= {\rm sgn} \big( \log K +  \log  p(M_1 ) - \log  p(M_0) \big) |\log K + \log  p(M_1 ) - \log  p(M_0 ) |^\frac{1}{\alpha}   \ \ .
\end{split}
\end{equation}
For equal model priors, $p(M_1 )=p(M_0 )$, we expect that
\begin{equation}
    \log  K  = \mathcal{J}_\alpha (f(^*x_O))  \ \ 
\end{equation}
\noindent for the optimal trained network $f^*$. 

As with many of the losses that we have previously described, the l-POP-Exponential Loss results in a (transformed) estimate of the log Bayes factor, which is ideal for Bayesian model comparison (e.g. the Jeffreys Scale).

We recommend a default value of $\alpha=2$.  This provides the reweighting of errors for high Bayes factor values, ensuring accuracy across many orders of magnitude, while being sufficiently simple that the network trains consistently well.

The l-POP transformation $\mathcal{J}_\alpha$ to act on $f(x)$ is not the only available choice that could be used to construct useful losses. Any strictly monotonic function $\mathcal{J}: \mathbb{R} \rightarrow  \mathbb{R}$ would fulfil the condition that $\mathcal{J}(f(^*x_O))=\log  K$. {Nevertheless, the choice is not arbitrary. For example, we have considered the $\rm sinh$ function in the place of $\mathcal{J}_\alpha$, but find the exponential scaling for high $K$ leads to unreliable performance. Since 
we have empirically found  our l-POP transform $\mathcal{J}_\alpha$ to give accurate results for several test cases presented in this paper we recommend it as the default choice.}. 

\begin{table}[ht]
\begin{center}
\normalsize
\begin{tabular}{||c c c||} 
 \hline
Loss name & $\mathcal{V}(f(x), m)$  &  $f^*(x_0)$  \\ \ & using model label $m \in \{0,1\}$  &  if $p(M_1)=p(M_0)$ \\ [0.5ex] 
 \hline\hline
Polynomial &  $m(1-f(x))^\alpha + (1-m) f(x)^\alpha$ &   $ p(M_1 | \ x_O) ^{\frac{1}{\alpha - 1}} $ \\ 
 \hline
Cross Entropy & $- m \log \big(f(x) \big)   - (1 - m) \log \big( 1 - f(x) \big) $ &   $ p(M_1 | \ x_O) $\\
 \hline
Exponential & $e^{(\frac{1}{2} - m )f(x)} $ & $\log K$\\ 
 \hline
Logistic & $\log \big( 1 + e^{(1-2m)f(x)} \big)$ & $\log K$\\
 \hline
$\alpha$-Exponential &  $\big( 1 + e^{(1 - 2m )f(x)} \big)^{\alpha-1}$  & $\frac{1}{\alpha}  \log K$ \\ 
 \hline
$\alpha$-log-Exponent & $ f(x)  ^{(\frac{1}{2}-m)\alpha}$ &  $K^{\frac{1}{\alpha}}$ \\
 \hline
l-POP-Exponential & $e^{(\frac{1}{2} - m )\mathcal{J}_\alpha (f(x))} $ & $\mathcal{J}_\alpha^{-1}(\log K)$  \\ 
 \hline

\end{tabular}
\end{center} 
\caption{A set of simple loss functions that can be used to estimate useful functions of the model posterior probability (in particular the Bayes factor $K$ for Bayesian model comparison).\label{tab:losses}}
\end{table}
\subsection{Summary of ``symmetric losses''} \label{sec:summary_symmetric}

In Table~\ref{tab:losses}, we summarize both the losses presented in the previous section and the corresponding optimal network outputs (the latter shown in terms of model posteriors or Bayes factors). 

This list is not at all comprehensive; we have not attempted to study the family of all loss functions that can be used to construct an Evidence Network. {Nor are all of these loss functions novel to this paper. The Squared Loss (a special case of the Polynomial), Cross-Entropy, Exponential, and Logistic have long been known in terms of Bayes risk for classification~\citep{loss_classification}. The ``$\alpha-$''type losses are new variations on existing losses and the POP or l-POP Exponential losses are novel to this paper. The main aim, however, is to use these losses in such a way that we can perform quantitative Bayesian model comparison via an Evidence Network.}

Furthermore, we have included only ``symmetric losses'', i.e. those that are symmetric with respect to relabelling $M_0$ and $M_1$. There are many alternative loss functions without this property; such functions  might be useful for certain applications (e.g. when it is known \textit{a priori} that the data will prefer one model far more than another). 

For a network to calculate $p(M_1|x)$, such an ``asymmetric'' loss might have the form:
\begin{equation}
    \mathcal{V}(f(x), m) = m \mathcal{A} (f(x)) + (1-m) \mathcal{B} (f(x)) \ \ ,
\end{equation}
\noindent for a choice of functions $\mathcal{A}$ and $\mathcal{B}$ where $\mathcal{A} (f(x))  \neq  \mathcal{B} (1 - f(x))$. {Some of these losses have been explored for computing likelihood-ratios using classifiers in~\cite{rizvi2023learning}. }

\section{Pitfalls \& validation techniques}
\label{sec:pitfalls}

\subsection{Challenge} \label{sec:pitfall_challenge}

Even if \textit{theoretically} there exists an optimal function $f^*$ that minimizes the global optimization objective equation~\eqref{eq:objective}, in practice three factors (typical of neural methods) prevent us from finding it exactly. First, we must approximate the space of functions with a set of neural network architectures. Second, we have access to only a finite number of samples from $p(x)$ and hence we only have a (Monte Carlo) approximation to the optimization objective, via:
\begin{equation}
    I[f] \approx  \frac{1}{N} \sum_{j=1}^N  \ \mathcal{V}(f(x_j), m_j) \ \ . 
\end{equation}
\noindent Third, we have no means of guaranteeing that a local optimum discovered during training is in fact a global optimum. It is, therefore, inevitable that there will be some error associated with our estimate of the Bayes factor.

In the next section (\ref{sec:timeseries}), we will demonstrate that the Evidence Network does indeed recover robust estimates of the Bayes factor.  {For such synthetic tests, we can fully validate our approach by choosing a model for which the true Bayes factors can be calculated analytically (i.e. in closed-form). 

{In actual applications, the likelihood and or prior are intractable or implicitly defined only through labelled training data  ({e.g.} consisting only of data realizations and their associated model labels from simulations or a generative model). In this setting a closed-form solution will  neither be available nor necessary since the true Bayes factors or model posteriors are not required to train an Evidence Network. Despite this, it is still necessary to validate the trained network. We will now describe a method for validating the Bayes factor computation by the Evidence Network that is generally applicable and requires only labelled training data of the same type that is needed to train the Evidence Network itself.}


\subsection{Blind coverage testing}  \label{sec:pitfall_solutions}

The solution to this unique validation problem is blind coverage testing. We can compare the estimate of $p(M_1 | x)$ to the fraction of validation data with the model label $M_1$.

Let us assume that the trained neural network is optimal, such that the network recovers the function $f^*
$. For example, if the network is trained using the l-POP-Exponential Loss (equation~\eqref{eq:popexploss}), for equal model priors we expect that:
\begin{equation}
    \mathcal{J}_\alpha (f^*(x_O))  = \ \log  K \ \ .
\end{equation}
For simplicity, consider the case of only two models $M_1$ and $M_0$ (this can be easily generalized for any number of models). We can then rearrange the previous expression to give the model posterior for $M_1$:
\begin{equation} \label{eq:rearrange}
    p(M_1 | \ x) = \frac{\exp \big(\mathcal{J}_\alpha (f^*(x_O))  \big)}{1+ \exp \big(\mathcal{J}_\alpha (f^*(x_O))  \big)}  \ \ .
\end{equation}
\noindent Such an expression can be constructed for any of the losses in Table~\ref{tab:losses} that estimate some function of $K$.

\newpage

Fix $P$ and let $\epsilon > 0$. Consider validation data $x^{\rm v} \sim p(x)$ for which the posterior probability $p(M_1 |x^{\rm v})$ lies in $\left[ P, P +\epsilon \right]$. As $\epsilon \rightarrow 0$, the expected fraction of such data having label $M_1$ converges to $P$. By comparing the true fraction of model labels in some small probability range with that expected from the estimated $K$, we can validate the trained network without knowing the true model posterior probabilities.  

{Put another way, consider a set of $N$ validation data samples that all have the same posterior probability $p(M_1 |x)$. In this set, if the number of data samples that are drawn from model $M_1$ is $n_{1}$, then the expected fraction of data from model $M_1$ is given by  $\langle \frac{n_{1}}{N} \rangle  = p(M_1 |x)$, following the usual definition of posterior probability. In practice, no two samples in the validation data have exactly equal posterior probabilities, so we use small bins of probability to calculate this fraction (corresponding to a small $\epsilon$). To quantify the statistical significance of the scatter in the probability bins, we can use a binomial distribution to model the count statistics (see section~\ref{sec:des} for an example).}

This procedure resembles the re-calibration of class probabilities in classification tasks~\citep{calibration1, calibration2}. In this work, it is used as a blind validation of our Bayes factor estimates for Bayesian model comparison. 

\section{Demonstration: Analytic time-series} \label{sec:timeseries}

\begin{figure}[ht]
\vskip 0.2in
\begin{center}
\centerline{\includegraphics[width=8cm]{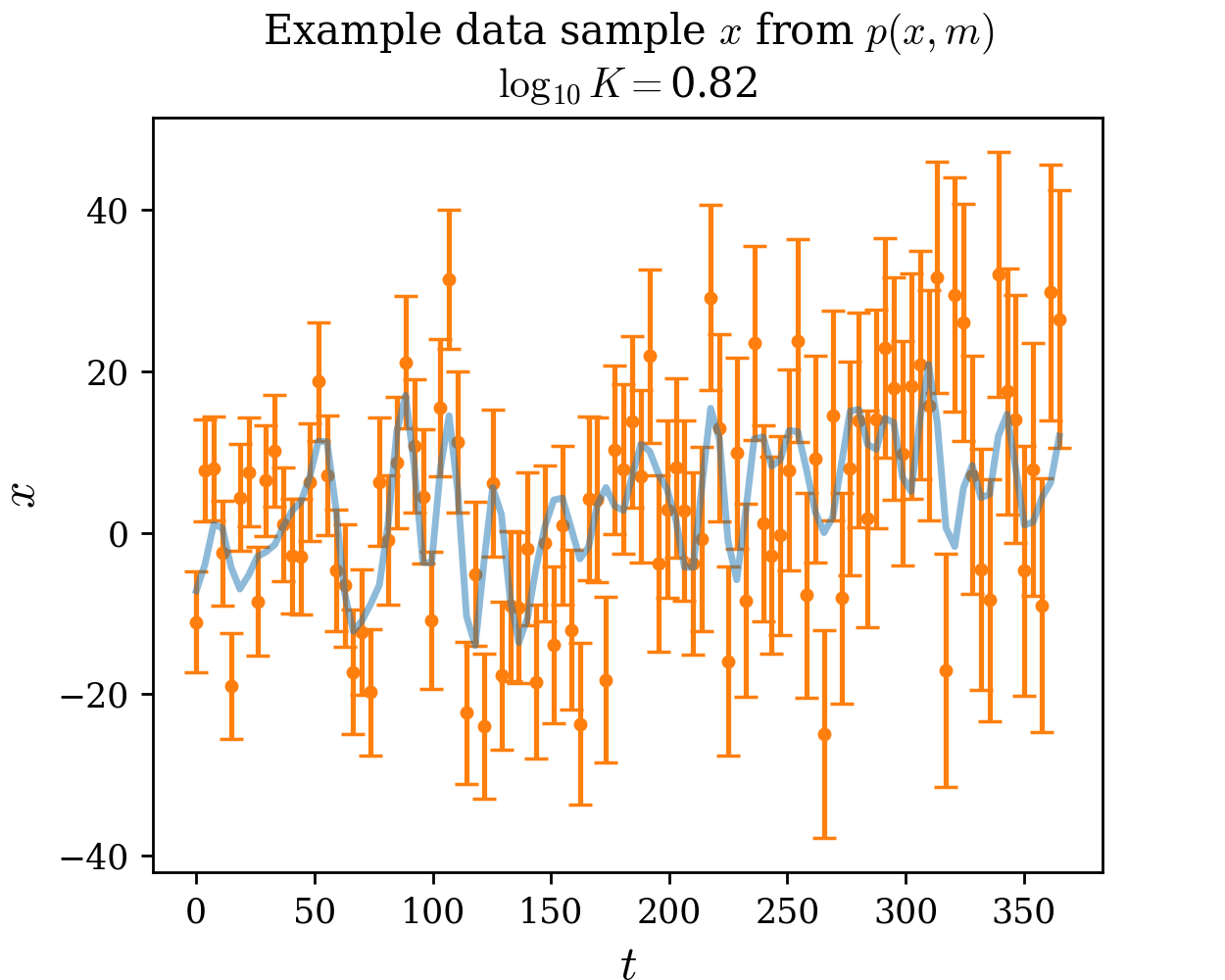}}
\caption{Example {100-parameter} time series model showing the underlying true signal overlaid with the observed data. The generative model is defined such that the Bayes factor can be calculated analytically from a closed form expression -- this is to {evaluate} the Evidence Network output {against} the {ground} truth for this demonstration. For this realization, $\log K > 0$, so $\theta_0=0$ is (slightly) disfavoured.}
\label{fig:time_series}
\end{center}
\vskip -0.2in
\end{figure}

\subsection{Overview}

We construct two generative models to sample time series data $x$. These are defined such that the model evidences and associated Bayes factors can be calculated analytically using a closed-form expression. In section~\ref{sec:genmodel}, this is compared to the estimated log Bayes factors from our simple Evidence Network.

We compare the Evidence Network to likely alternative methods. Even if the true likelihood/prior were known, with the use of the most advanced nested sampling methods~\citep{polychord} the problem can still be too arduous for high parameter-space dimensionality $\mathrm{dim}(\theta) \gtrsim 10^2$; we demonstrate this in section~\ref{sec:polychord}. Assuming that we only have data samples and model labels, we would have no knowledge of the distributions $p(x | \theta)$ or $p(\theta)$ to evaluate the marginal likelihood (equation~\eqref{eq:marg_ev}). As previously discussed, one could try to estimate those probability densities using implicit inference, but with high data dimensionality $\mathrm{dim}(x) \gtrsim 10^2$ or high parameter-space dimensionality $\mathrm{dim}(\theta) \gtrsim 10^2$, density estimation becomes intractable or prone to significant error; we show this in section~\ref{sec:density_method}.

\begin{figure}
\begin{center}
\centerline{ \hspace{1cm} \includegraphics[width=18.cm]{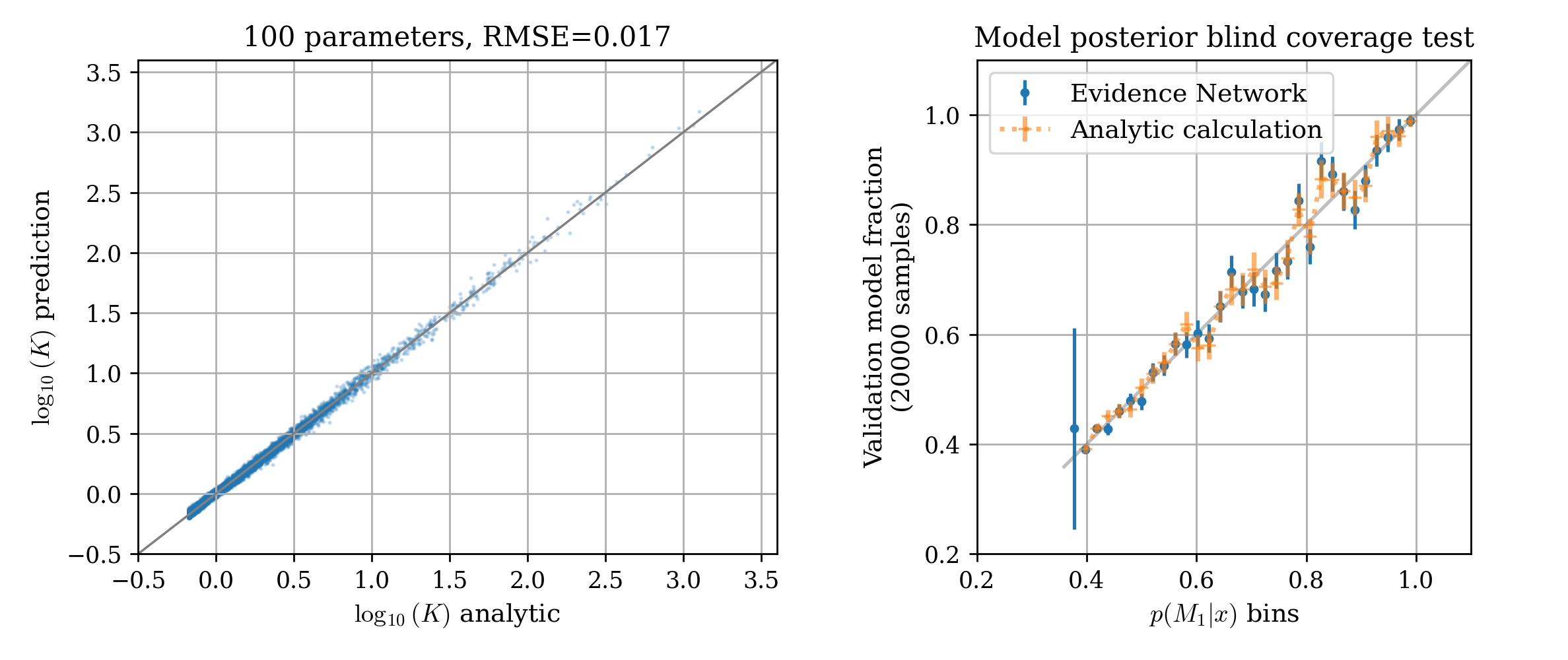}}
\vskip -0.2in
{\caption{\textit{Left panel}: Estimated log Bayes factor $K$ using an ensemble Evidence Network (with four networks) compared to analytic calculation using a closed-form expression for $K$. These data are time series with $100$ data elements draw from a generative model with $100$ model parameters. This result uses our default network architecture with $10^6$ training samples. As we show in section~\ref{sec:timeseries}, all standard methods to compute the Bayes factor either fail or incur significant error for this high-dimensional example. \textit{Right panel}: Evidence Network model posterior probabilities derived from the estimated Bayes factors (equation~\eqref{eq:rearrange}) compared with the relative model fraction in the validation data. This allows validating the Evidence Network output when ground-truth model evidence is unavailable.} \label{fig:100_gaussian} }
\end{center}
\vskip -0.2in
\end{figure}

\subsection{Generative model} \label{sec:genmodel}

We construct two nested models that generate time series data with heteroschedastic random noise. Both models are linear Gaussian models with a linear operation that is non-trivial with respect to time.

The prior for the model parameters $\theta$ is a simple normal distribution for each parameter $\theta_i$:
\begin{equation}
p(\theta_i) = \mathcal{N}(\theta; 0,1) \ \ .
\end{equation}
\noindent Per data (vector) realization $x$, each element $x_j$ is given by:
\begin{equation}
x_j = \sum\limits_{i} A_{ji} \theta_j + n_j  = \mu^x_j + n_j  \ \ ,
\end{equation}
\noindent where the heteroscedastic noise is drawn from a multivariate normal distribution, such that:
\begin{equation}
 p(n) = \mathcal{N} (n;0, {\Sigma}) \ \ .
\end{equation}
\noindent The noise covariance ${\Sigma}$ is a diagonal matrix with diagonal elements ${\Sigma}_{k k}$ given by:
\begin{equation}
 \sqrt{{\Sigma}_{k k}} = \sqrt{\frac{N}{100}} \Big( \big(k+\frac{5}{2} \big) \frac{8}{5}\Big)^2 \ \ .
\end{equation}
\noindent For the first model, $M_1$, the linear operation is given by:
\begin{equation}
A_{ji} = \cos \big((i-\frac{1}{2})t_j \big)
\end{equation}
\noindent for $i>0$. For $i=0$, we have
\begin{equation}
\ A_{j0} = 2 t_j  \ \ .
\end{equation}
The $i=0$ term corresponds to a linear growth over time. For the second model, $M_0$, this first term is removed, so there is no linear growth term over time. This is equivalent to setting $\theta_0=0$.

In both of these cases we can evaluate the marginal evidence integral (equation~\eqref{eq:marg_ev}) analytically to recover a closed-form solution for the Bayesian model evidence, which is equivalent to evaluating a normal distribution:
\begin{equation}
p(x | \mathcal{M}) = Z(x| \mathcal{M}) = \mathcal{N}(x ; \mu_Z, C_Z) \ \ ,
\end{equation}
\noindent where
\begin{equation}
\mu_Z = A \cdot \mu^x \ \ ,
\end{equation}
\noindent and
\begin{equation}
C_Z = \Sigma +  A A^T \ \ .
\end{equation}
This can be simply evaluated for each model to calculate the ratio to give the Bayes factor $K$:
\begin{equation}
K = \frac{Z(x | M_1)}{Z(x | M_0)} \ \ .
\end{equation}

\begin{figure}[ht]
\begin{center}
\centerline{\includegraphics[width=17.2cm]{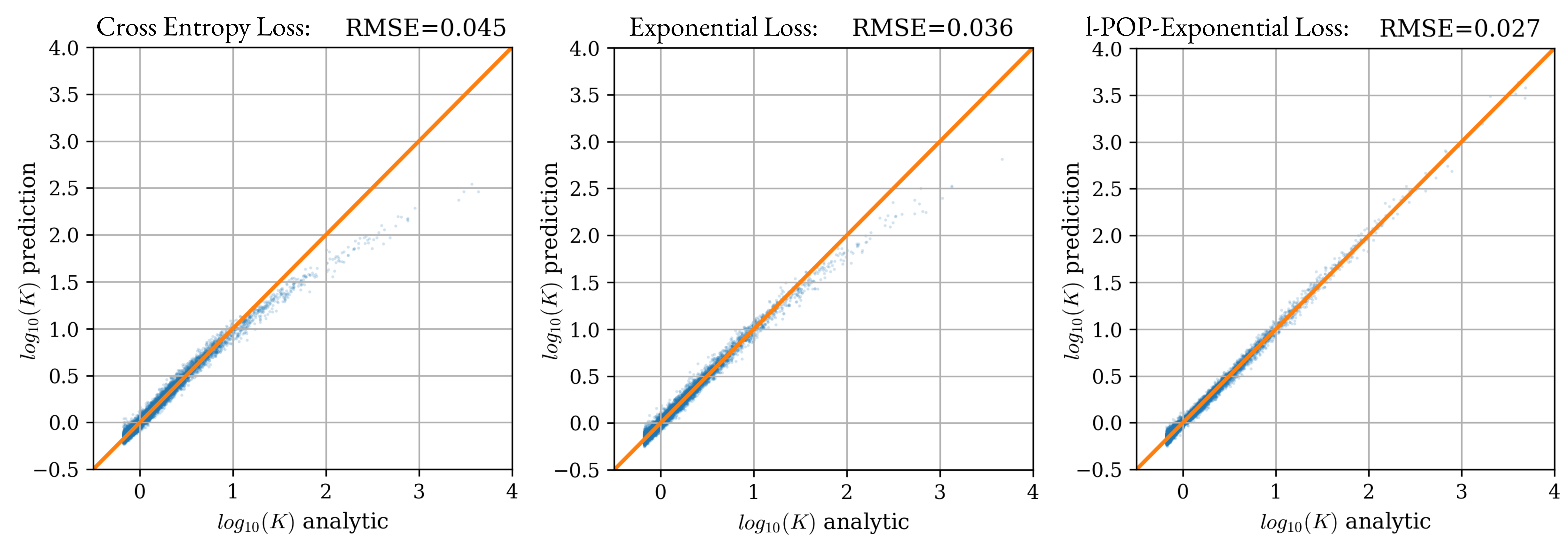}}
\vskip -0.1in
{\caption{Three choices of loss to estimate the log Bayes factor $K$. Each panel shows results from the same network architecture; we use only a single network, rather than an ensemble as in Fig.~\ref{fig:100_gaussian}, as this is sufficient to show the different systematic errors with increasing $K$. The \textit{left panel} uses the Cross Entropy Loss to estimate individual model posteriors from which the Bayes factor is estimated. The \textit{centre panel} uses the Exponential Loss to estimate $\log K$. The \textit{right panel} uses the l-POP-Exponential Loss ($\alpha=2$) to estimate $\log K$.\label{fig:loss_comparison}}}
\end{center}
\vskip -0.2in
\end{figure}

\subsection{Evidence Network results} \label{sec:timeseries_results}

Using data drawn from the generative model described in the previous Subsection, we train the Evidence Network using the l-POP-Exponential Loss with $\alpha=2$ to recover a simple transformation of $\log K$ (see Table~\ref{tab:losses}).  This choice of loss function is our default recommendation:
\begin{equation}
{\rm l}\mhyphen{\rm POP}\mhyphen{\rm Exponential\ Loss: \ } \ e^{\mathcal{J}_{\alpha=2}(f(x))} = e^{(\frac{1}{2} - m )(f(x)+f(x)|f(x)|)}  \implies f^*(x_O)(1  +  |f^*(x_O)|)= \log K 
\end{equation}
\noindent Using the model labelling described above, a Bayes factor greater than one corresponds to a preference for model $M_1$ (i.e. $\theta_0=0$ is not preferred in this case).

We find that taking the average output of an ensemble of networks, each with independent random weight initialization, significantly improves the accuracy of our log Bayes factor estimation. For these results, each network has the identical architecture.

We have chosen a simple network with little tuning that uses 6 dense layers (see~\ref{append:network} for details). The network was purposefully chosen to be simple to demonstrate that the network does not need to be tuned for the Bayes factor estimates to be reliable.

{Though we have chosen a simple dense architecture, Evidence Networks can use any type of network architecture, provided they use the correct optimization objective and are validated using our proposed coverage test. Depending on the application, they could use geometric architectures (e.g. deep sets, graph networks; \citealt{battaglia2018relational}), recurrent neural networks (e.g. long short-term memory --- LSTM; \citealt{lstm}), or any appropriate architecture.
}

Our main result uses the generative model described above with $\mathrm{dim}(\theta) = 10^2$ and $\mathrm{dim}(x) = 10^2$.  As the data samples for this generative model are symmetric under a sign flip, we make use of this standard data augmentation during training. 

The left panel of Fig.\ref{fig:100_gaussian} shows our main result with the ensemble Evidence Network. We recover estimates of the Bayes factor over many orders of magnitude. When compared with the analytic (true) Bayes factors, we calculate the root-mean-square error (RMSE) and find a $\log K$ error $\approx 0.02$. 


This is an excellent estimation of the Bayes factor for such a high dimensional problem. The high dimensionality of the data and parameters render this task either intractable or error-prone for alternative methods as we demonstrate in the following subsections.

For most applications, the true Bayes factor value will typically not be available for validation; our training data consists only of data realizations with their associated model label. We can, however, leverage our blind coverage test to validate the model posterior probabilities are recovered. This is shown in the right panel of Fig.~\ref{fig:100_gaussian} for the same trained network and data.

The plotted error-bars in the right panel of Fig.~\ref{fig:100_gaussian} are the expected standard error for binomial distributed samples: $\sigma_{\rm err} = \sqrt{p(M_1|x_O) (1 - p(M_1|x_O))/n_p}$, where $n_p$ are the number of data samples in a given $p$ probability bin. We find the measured standard deviation of the residual error (between the expected and the predicted probabilities) does indeed match $\sigma_{\rm err}$,
further validating the method. 

For a general problem where the true Bayes factors are not known, the result of such a blind coverage test can be used to construct error bounds for the Bayes factors estimates.

Fig.~\ref{fig:loss_comparison} shows different choice of loss functions. The \textit{left panel} does not use a loss function to estimate the Bayes factor directly, but uses the Cross Entropy Loss to individually estimate $p(M_1 | x_O)$ and $p(M_0 | x_O)$, whose ratio gives the Bayes factor $K$. This approach leads to systematic inaccuracies for moderately high values of $K$. These inaccuracies are ameliorated with the choice of the Exponential Loss (\textit{centre panel}), from which the Evidence Network directly estimates $\log K$. Using the l-POP-Exponential Loss (\textit{right panel}) the systematic error at high $K$ is no longer apparent.

\begin{figure}[ht]
\begin{center}
\centerline{\includegraphics[width=16.2cm]{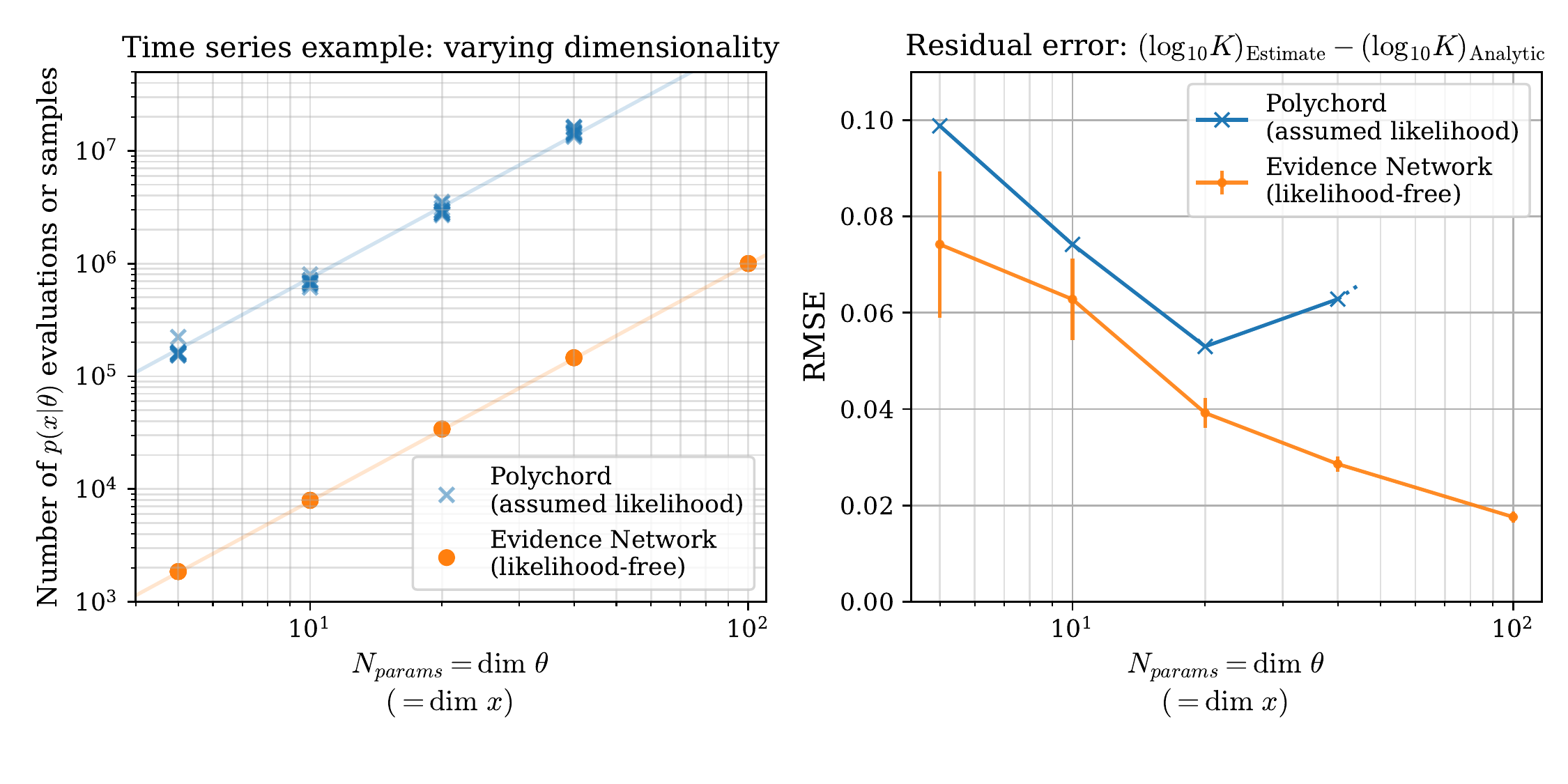}}
\vskip -0.4in
{\caption{\textit{Left panel:} The number of nested sampling $p(x|\theta)$ evaluations is controlled by the number of live points in the PolyChord algorithm, which used the code default value (with a maximum cut-off of $10^8$). The number of $p(x|\theta)$ samples used by the Evidence Network {(i.e. $x$ samples for training)} was chosen to match the polynomial scaling of PolyChord (solid lines), but fixed at approximately $\sim 1$ per cent of the number of PolyChord evaluations {of $p(x|\theta)$}. \textit{Right panel:} Despite significantly fewer $p(x|\theta)$ samples than nested sampling $p(x|\theta)$ evaluations, we consistently recover more accurate estimates of $\log K$ with the Evidence Network. {The Evidence Network error-bars are the standard deviation of the} \textsc{rmse} {result from 5 runs of the Evidence Network ensemble (each with four networks); the central value is the mean. For the PolyChord points, it is infeasible to estimate an error-bar given the computational time required.}}}
\label{fig:polychord_scaling}
\end{center}
\vskip -0.2in
\end{figure}
\subsection{Comparison to state-of-the-art: nested sampling with known likelihood} \label{sec:polychord}

In this subsection we compare the Evidence Network with nested sampling; we assume the correct likelihood for nested sampling, while restricting the Evidence Network to the likelihood-free problem, and find that the Evidence Network is still able to outperform nested sampling for high-dimensional parameter spaces. 

Let us assume we know the likelihoods, $p(x | \theta, M_1)$ and $p(x | \theta, M_0)$, and priors, $p( \theta |  M_1$ and $p( \theta | M_0)$. We assume that we are able to either evaluate a given likelihood, $\mathcal{L}_1(\theta)=p(x_O | \theta, M_1)$,
or sample data realizations from it, $x_i \sim p(x | \theta, M_1)$. For a Gaussian likelihood, these two operations are of equal computational complexity.   

We use the PolyChord~\citep{polychord} code to perform the nested sampling; this uses an advanced algorithm that combines both cluster detection for multi-modal distributions and slice sampling using affine ``whitening'' transformations. We use the default number of {\sc live points}$=25\times {\rm dim} (\theta)$, which then leads to a certain number of $p(x|\theta)$ evaluations. We use a maximum cut-off of $10^8$ likelihood evaluations for reasons of computation time.

We choose the number of $p(x|\theta)$ samples for the Evidence Network to match the same polynomial scaling with ${\rm dim} (\theta)$ as PolyChord evaluation, but with only approximately $1$ per cent of that of PolyChord. Despite significantly fewer $p(x|\theta)$ samples than nested sampling $p(x|\theta)$ evaluations (Fig.~\ref{fig:polychord_scaling}, \textit{left panel}), we consistently recover more accurate estimates of $\log K$ with the Evidence Network (Fig.~\ref{fig:polychord_scaling}, \textit{right panel}). 

{This is an indication that it may be more efficient to train an Evidence Network to estimate Bayes Factors \textit{even when the likelihood is known}}. For many closed-form likelihoods, it is a relatively simple code change to replace the $p(x|\theta)$ evaluation with a $p(x|\theta)$ sample of $x$.

\subsection{{Alternative likelihood-free method:   density estimation}} \label{sec:density_method}

Let us now consider alternative approaches that are available for likelihood-free (e.g. simulation-based) inference problems.

As previously discussed in section~\ref{sec:compare}, we could first learn the form of $p(x | \theta ) $ from the training data; this is known as Neural Likelihood Estimation (NLE). We would still need to perform the high-dimensional marginal evidence integration and using nested sampling (or a similar method). This step is unnecessary: the intermediate density estimation step can only induce potential errors and we nevertheless return to the difficult procedure discussed in section~\ref{sec:polychord}.

An alternative density estimation approach would be to learn the form of $p(x | \mathcal{M})$ from the training data. So far, this method has not been widely considered for scientific model comparison, even though it would also avoid marginal evidence integration. Therefore, we compare Evidence Networks with this density estimation approach.

We use an ensemble of neural density estimators to estimate $p(x | M_1)$ and $p(x | M_0)$ and use an ensemble of neural networks (matching the set-up used so far) to form the Evidence Network to estimate $\log K$. In both cases we use $10^6$ samples of training data $x$. Density estimation suffers particularly badly from increasing dimensionality. We will, therefore, reduce from a dimensionality of ${\rm dim} ( x ) = 100$ (as in Fig.~\ref{fig:100_gaussian}) to a more manageable ${\rm dim} (x) ={\rm dim} (\theta) =  20$ for this comparison.

We use Normalizing Flows for neural density estimation; we use Neural Spline Flows~\citep{neuralsplines}, which are a popular and flexible neural density estimation method. As an implementation, we use the 
\texttt{PZFlow} package~\citep{pzflow} with the default settings.

Fig.~\ref{fig:norm_flow} shows the comparison between the two methods. The {\textit left panel} shows the analytic $\log K$ value against the estimated value using both the Evidence Network and the Normalizing flow ratio (using $p(x | M_1)/p(x | M_0)$). The relatively large residual scatter for the Normalizing Flow can also be seen in the {\textit{right panel}}, which shows the histogram of the residual errors.

\begin{figure}[ht]
\begin{center}
\centerline{\includegraphics[width=15.2cm]{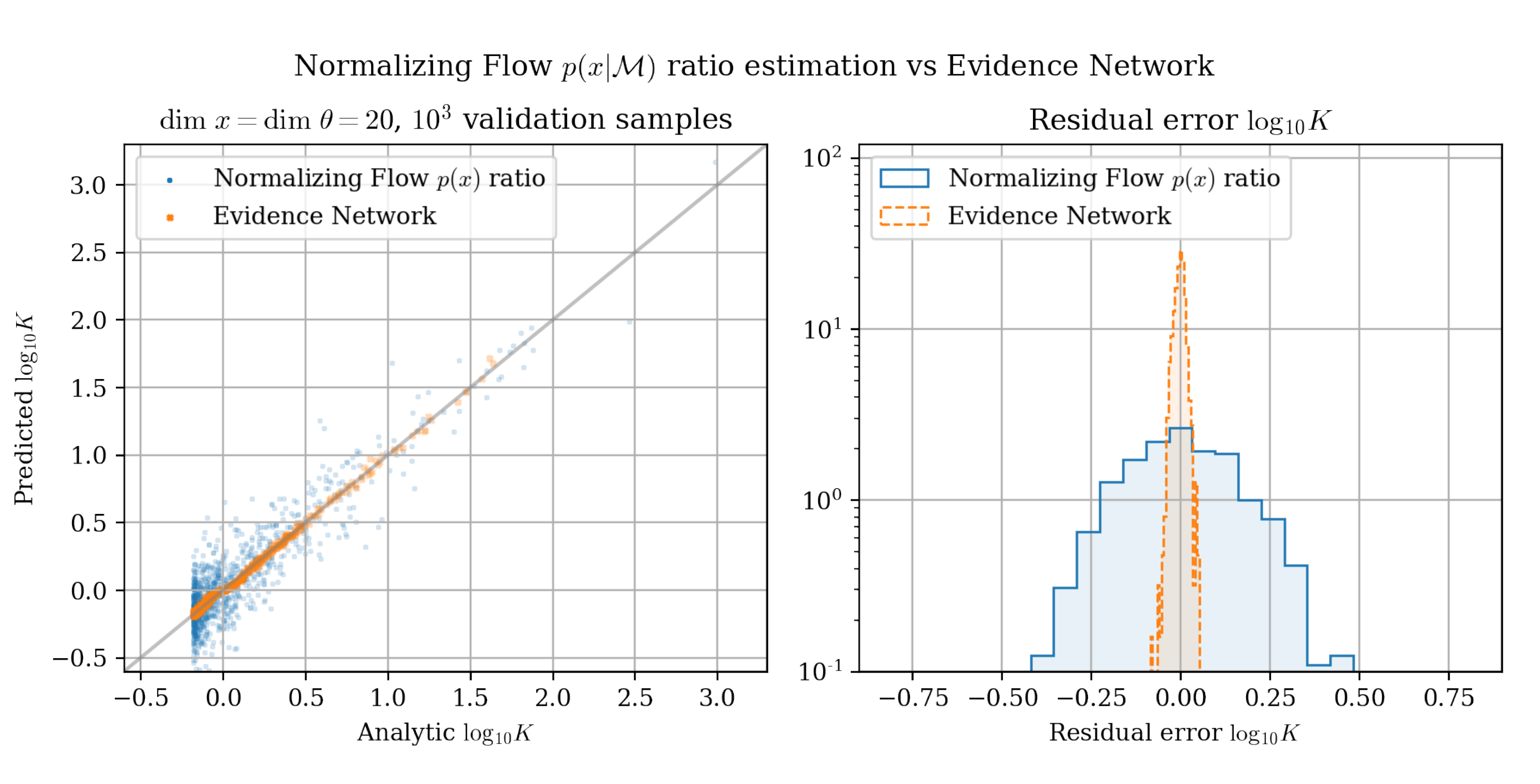}}
\vskip -0.2in
{\caption{\textit{Left panel}: Estimated $\log_{10} K$ for 1000 data samples using (i) the Normalizing Flow ratio for estimated $p(x|M_1)$  and $p(x|M_0)$, and (ii) the direct estimate using the Evidence Network. The set-up for this prediction using the 20 dimensional time series data is described in section~\ref{sec:density_method}. \textit{Right panel}: The distribution of residual errors for each method. The RMSE (error) using the Normalizing Flow is more than a factor of 10 larger than the direct estimation of $\log_{10} K$ using the Evidence Network for this problem.}
\label{fig:norm_flow}}
\end{center}
\vskip -0.2in
\end{figure}

\begin{figure}[ht]
\begin{center}
\centerline{\includegraphics[width=17.5cm]{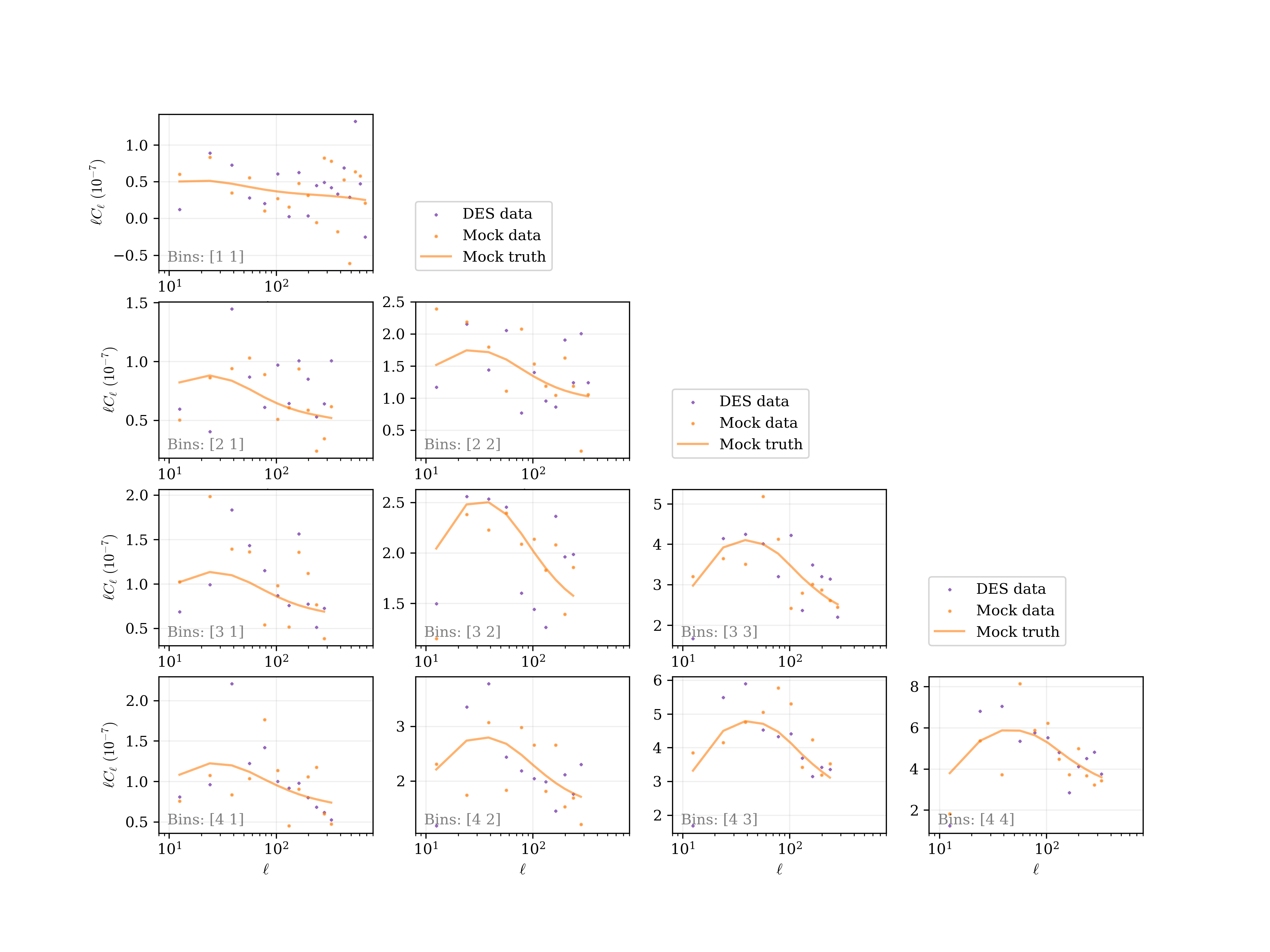}} 
\vskip -0.4in
{\caption{Dark Energy Survey Year 3 weak gravitational lensing power spectra and cross-spectra. Each panel has a ``Bins'' label corresponding to the pair of tomographic (redshift) bins for each spectra. The purple squared markers are the observed Dark Energy Survey data $x_O$. The orange circular markers are mock data $x_i$ generated from our model, for which the orange solid line is the theoretical, expected signal without noise. 
\label{fig:des_data}} }
\end{center}
\vskip -0.2in
\end{figure}

The RMSE (error) in the $\log K$ using the Normalizing Flow is more than a factor of 10 larger than the Evidence Network. The reason for this is twofold:  (i) density estimation suffers exponentially from increasing dimensionality (which is even a problem for this restricted 20 dimensional data), and (ii) the errors in each $p(x | \mathcal{M})$ combine in the ratio that gives $\log K$.

\section{Demonstration: Dark Energy Survey data} \label{sec:des}

{
The Dark Energy Survey (DES) is a major, ongoing galaxy survey that, among other science goals, measures the shapes of galaxies in order to use the gravitational lensing effect to make cosmological inferences.  With its wide-field camera mounted on the Blanco 4-meter telescope in Chile, DES captures high-resolution images of the southern sky, allowing for the measurement of lensing distortions using the shapes of millions of galaxies. By analysing these distortions statistically, DES provides  information on the distribution of matter, both visible and dark, on large scales. However, to interpret these observations accurately, it is crucial to account for intrinsic alignment of the galaxies, as this can introduce systematic biases. For a summary of weak gravitational lensing with the Dark Energy Survey Year 3 data see~\cite{amon, secco, mass_map}. In this section, we demonstrate the use of Evidence Networks to diagnose or detect intrinsic alignment effects. The aim of this analysis is to demonstrate the ease of Bayesian model comparison with Evidence Networks for real-world data.}

Using DES as a practical example, we construct a simple Bayesian model comparison problem. One model, $M_1$, can be summarized as: galaxies shapes are intrinsically aligned according to the simple non-linear alignment (NLA) model with an unknown amplitude $A_{IA}$ with prior probability $p(A_{IA}) = \mathcal{U}(-3,3)$. In the other model, $M_0$, galaxies have no alignment, i.e. $A_{IA}=0$.

The data for our test are the auto- and cross-spectra (angular power spectra) of the weak gravitational lensing signal in tomographic bins at varying distances (characterized by the redshift of galaxies in each bin). We use the same data, with same data cuts of small angular scales, as described in~\cite{doux}. The data model is the $\Lambda$-cold-dark-matter ($\Lambda$CDM) model, with a Gaussian likelihood and priors for cosmological parameters matching~\cite{doux} and all others unknown systematic parameters fixed to their fiducial values for simplicity. 

{To generate the training data under the assumed Gaussian likelihood, we simply draw data realizations from $p( x |  \theta, \mathcal{M})$,  in which the model parameters $\theta$ are themselves random draws from the prior $p(\theta | \mathcal{M})$. In practice, this is a simple change in the likelihood and prior code to sample from each probability distribution instead of evaluating each probability distribution.}

\begin{figure}[ht]
\vskip 0.2in
\begin{center}
\centerline{\includegraphics[width=16.cm]{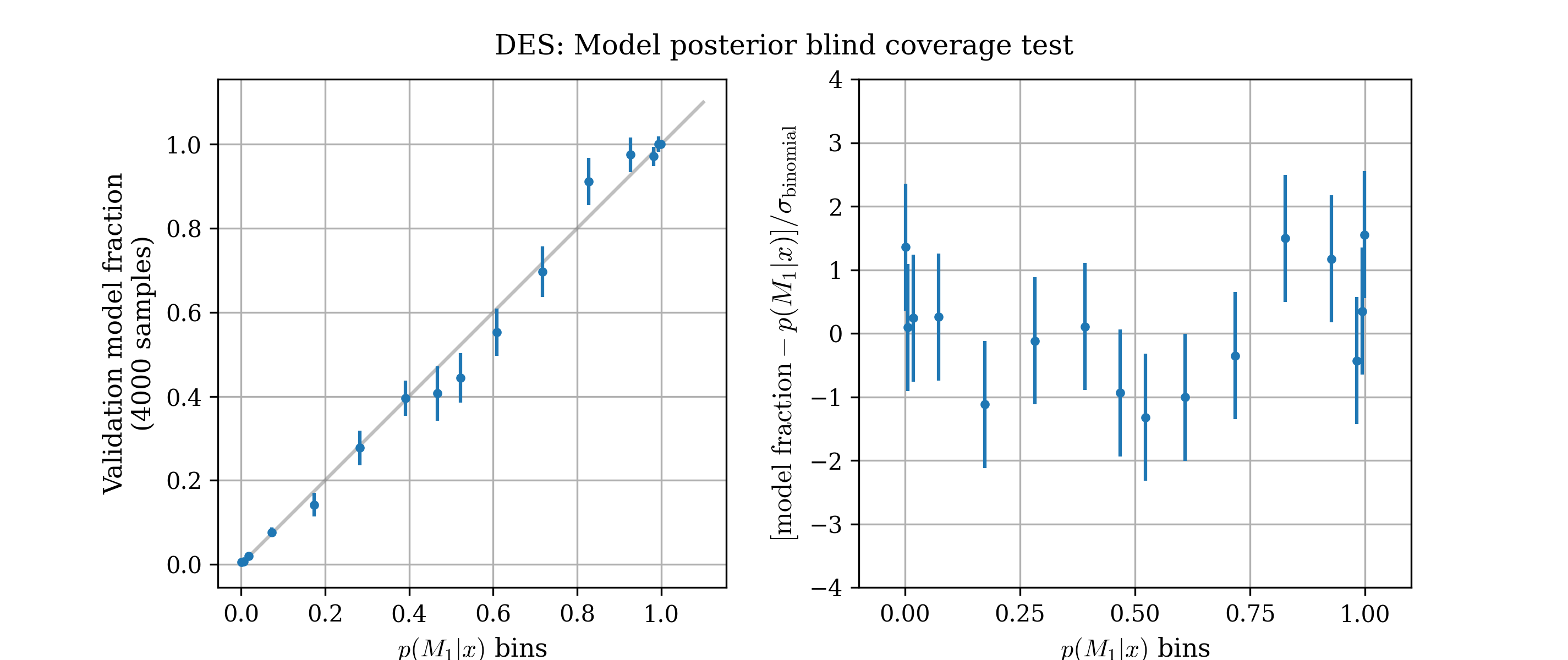}}
\caption{Blind coverage test of trained Evidence Network for the Dark Energy Survey (DES) weak gravitational lensing data (section~\ref{sec:des}). \textit{Left panel:} Evidence Network model
posterior probabilities derived from the estimated Bayes factors compared with the
relative model fraction in the validation data. The error bars are simple standard deviation values for a binomial distribution. \textit{Right panel:} The residual between the model fraction and the posterior probability from the Evidence Network, which has been rescaled by the expected standard deviation (for binomial distributed data). The scatter around zero matches the expected sample variance, which validates the Evidence Network. }
\label{fig:des_coverage_test}
\end{center}
\vskip -0.2in
\end{figure}

Fig.~\ref{fig:des_data} shows the DES data in the form of the angular power spectra, $C_\ell$. In addition to the observed data $x_O$, we show mock generated data (including noise) along with the mock truth (the theoretical expectation). We generate $2\times 10^4$ data samples $x_i$ from each model using {a} Gaussian likelihood assumption. We train an Evidence Network to estimate $\log K$, using an ensemble of 13 networks with the same architecture as in section~\ref{sec:timeseries}, but with one layer removed, a longer training time, and a higher initial learning rate ($10^{-3}$).

\textbf{We find $\alpha=1$ gives reliable performance for this example {in terms of consistent training. The default choice of $\alpha=2$ led to over-fitting for some networks of the ensemble, but this is simple to diagnose and solve with a different (typically lower) choice of $\alpha$. Poor coverage test results can also be used to decide to use a non-standard value of $\alpha \neq 2$. These decisions are described in \ref{append:network}.}}

For this problem we do not have the analytic Bayes factor calculation available for each element in our validation data (as we did in section~\ref{sec:timeseries}), so we take advantage of our blind coverage test (described in section~\ref{sec:pitfalls}). Fig.~\ref{fig:des_coverage_test} shows this coverage test. The {\textit{right panel}} shows the residuals between the model fraction and the expected $p(M_1 |x)$ derived from the Evidence Network. This has been rescaled by $\sigma_{\rm binomial}$, the expected standard deviation for the binomial distribution. The scatter (around zero) of the rescaled residuals is what we would expect given $\sigma_{\rm binomial}$.

Evaluating our trained Evidence Network on the observed DES weak gravitational lensing (power spectrum) data, we find: $$\log_{10} K(x_O)=-0.8 \ (\pm 0.3) \ .$$ This gives a very mild preference for the model without intrinsic alignments of galaxies with odds of approximately $6/1$. The quoted error is a simple jacknife resampling estimate using the 13 individual estimates from the network ensemble.

This result is not meant as a novel scientific result (not least because the systematic error modelling has been simplified), but it serves to show the simple steps involved in such a Bayesian model comparison with Evidence Networks.

\section{Pedagogical Example: Rastrigin Posterior} \label{sec:rastrigin}

We have discussed the advantages of Evidence Networks in cases where the likelihood and/or the prior are not explicitly known but only implicit in the form of simulated data samples with model labels.
In this section, we aim to address a separate conceptual point: the possibility of straightforward model comparison  in cases where computing the posterior for inference is intractably hard, even when prior and likelihood are explicitly specified up to parameter-independent normalisation constants. To illustrate this concept, we present a synthetic data case study involving a highly multimodal posterior density. We compare two models: one model where the components of the parameter vector $\theta_i$, $i=1,\dots,n$ have a Rastrigin prior with well-separated modes, 
\begin{equation}
    p(\theta_i|M_1)\propto e^{-\frac{\theta_i^2}{4}+10 (\cos(2 \pi  \theta_i)-1)}, \quad{\mathit{(Rastrigin)}}
\end{equation}
and a second model that omits the oscillatory term in the exponent, leading to the Gaussian prior 
\begin{equation}
    p(\theta_i|M_0)\propto e^{-\frac{\theta_i^2}{4}}. \quad{\mathit{(Gaussian)}}
\end{equation}
We choose the Rastrigin function as it is commonly used to illustrate the challenges of multimodality for optimization and inference problems, e.g. as in \citealt{polychord}. 

Consider the data model $x=\theta+n$ where $n$ is normal, $n\sim N(0,1)$. Taking the data $x=0$ for illustration results in the highly multimodal posterior  shown in Figure \ref{fig:rastriginPosterior} for the $n=2$ dimensional case.   Classical approaches to model comparison such as annealed importance sampling, thermodynamic integration, or nested sampling are methods to integrate the sampled posterior. The Rastrigin model produces a highly multimodal posterior whose number of well-separated modes increases exponentially with the number of parameters $n$, causing classical methods such as nested sampling to require an exponentially large number of samples in order to succeed. Similarly, approximating $p(x|M_i)$ with a  neural density estimator such as a normalizing flow is generally challenging for large $n$ since the absolute normalization depends on the global properties of the distribution.

This problem is most severe when  the data are noisy, since the posterior will be similar to the prior, and hence contain a large number of modes for each parameter dimension. We will show that this exponential complexity in the posterior can in fact be entirely avoided in our approach, since the evidence or Bayes factor make no reference to the parameters.  

In fact, focusing on the Bayes factor, the computation becomes trivial in the noisy limit. The easiest way to see this conceptually is that for sufficiently large noise, the probability densities of the data $p(x|M_0)$ and $p(x|M_1)$ become identical, and the Bayes factor will approach unity because the models are indistinguishable. For large noise, the prior is the same as the posterior and the Bayes factor is approximately 1 across all data realisations generated under the two models. 

The Evidence Network trivally does the right thing. It  simply ``sees'' indistinguishable data simulations from both models and will learn  to predict 1 regardless of input.

We show a numerical example for a non-trivial case with  moderate  noise variance set to 1/16 in Figure \ref{fig:rastriginPosterior}. Even in this case the Bayes factor oscillates around 1 with an amplitude of $O(1)$. The Evidence Network  learns the local relative probability for any particular data set to be drawn from one model or the other, correctly estimating the Bayes factor for the range of data that is represented in the training set.

The lesson from this example is that there are cases where estimation on the Bayes factor for model comparison (e.g. with Evidence Networks) can be done even in cases where solving the parameter inference problem is intractable (thus prohibiting nested sampling or annealing). Note that we do not claim that Evidence Networks can always compute the Bayes factor trivially\footnote{For example, in the large $n$ limit of our current example,  the Bayes factor would become a very rapidly varying function of the data and would require large amounts of training data and a highly expressive neural network to converge.}. Our claim is that Evidence Networks can provide accurate estimates of the Bayes factor in regimes where other methods (and notably nested sampling) fail due to the high complexity of the posterior. 

\begin{figure}[ht]
\vskip 0.2in
\begin{center}
\centerline{\includegraphics[width=16cm]{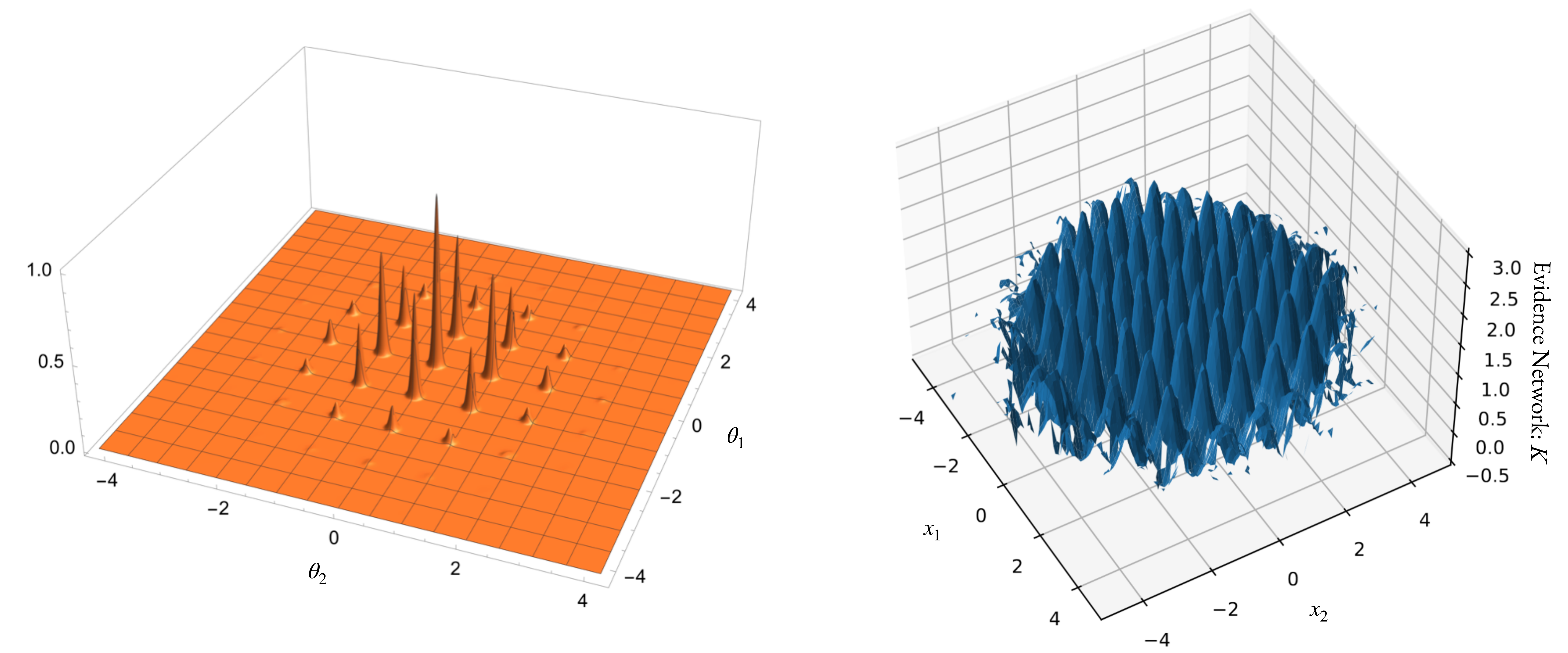}}
\caption{{\textit{Left panel:}} Marginal posterior for two parameters of the Rastrigin model {(defined in section~\ref{sec:rastrigin})}. \textit{Right panel:} The Bayes factor $K$, as a function of two data elements $x_1$ and $x_2$, estimated using an Evidence Network evaluated on the validation data.}
\label{fig:rastriginPosterior}
\end{center}
\vskip -0.2in
\end{figure}

\section{Conclusion and extensions} \label{sec:conclusion}
Computing the the Bayes factor (evidence ratio) for  Bayesian model comparison presents a significant challenge in computational statistics. The leading approaches, such as nested sampling, can require the evaluation of intractable integrals over the space of permissible model parameters. This problem becomes increasingly difficult as the dimensionality of the parameter space or the complexity of the posterior increases. We have introduced Evidence Networks as a solution to this problem. We cast the computation as an optimization problem, solved by neural networks trained on data realizations from the model with a designer loss. Couched in this way, the computation makes no reference to the parameters of the underlying model, does not require exploring the posterior density, and  can use uncompressed data. The result of the optimization is an amortized neural estimator of the Bayes factor that can be applied to any data set in a single forward pass through the neural network. 

In addition to decoupling the model comparison problem from the unnormalized posterior or the dimensionality of the parameter space, our approach is also suited for implicit (likelihood-free or simulation-based) inference, where the explicit forms of either the likelihood and prior distributions are unknown or intractable. 
All that is required to train our estimator are examples of data from the models that are to be compared. These would generally be computer simulations from the different models, they could be the result of neural generators or be labelled real-world data sets.

In accordance with expectation, we observe in numerical experiments that the estimator performance is robust to the dimension of parameter space. For the training set sizes and example problems we have studied the fractional accuracy of the computed Bayes factor is of order 2 per cent with accurate predictions over many orders of magnitude. Given the subjective nature of interpreting the strength of evidence for model selection, this is sufficient for many applications of model comparison.

We will now discuss a number of ways to use Evidence Networks straightforwardly for applications that go beyond the Bayes factor or model posterior probabilities.

\subsection{Simple extensions for further applications } \label{sec:extensions}
\paragraph{Absolute evidence.}
We have focused our discussion on model comparison based on (functions of) the evidence ratios. We compute these based  on (functions of) the posterior model probabilities, e.g. the log Bayes factor, since these are what our approach returns natively.

Should the absolute evidence be required, for model $M_1$, say, this can  be obtained easily: choose model $M_0$ to be a reference model where the evidence is analytically tractable (such as a linear Gaussian model), giving $p_\mathrm{analytic}(x|M_0)$; we can then use our method to compute the Bayes factor $K$ and then solve for the required evidence as $p(x|M_1)=K p_\mathrm{analytic}(x|M_0)$.

\paragraph{Connection to frequentist hypothesis testing.}
Numerous writers have advocated the Bayesian formalism as a way to generate statistical procedures with good frequentist properties, regardless of the philosophical differences between the two approaches  (see for example \citet{carlin2010bayes}). In the context of hypothesis testing this has taken the form of deriving practical guidelines for frequentist test thresholds ($p$-values) that are calibrated on the  Bayes factor for certain applications \citep{Johnson2013RevisedSF}. To make contact between Bayes factors and significance thresholds more broadly it would be helpful to be able to study the frequentist properties of the Bayes factor. The computational challenges that motivated our present study have so far prevented this as the Bayes factor for each data sample would be necessary. But Evidence Networks are amortized, meaning it is near-instantaneous to  compute accurate Bayes factors for simulated data sets. This permits the use of Bayes factors as a test statistic in hypothesis testing for a broad set of problem classes, including those not covered by existing guidelines.

\paragraph{Using Evidence Networks to do posterior predictive tests.}
While the interpretation of the Bayes factor is clear, the strong dependence on prior choice can be seen as a weakness. A common alternative is posterior predictive testing (PPT, see~\citealt{gelman2013bayesian} for overview). PPT partitions the data $x$ into subsets,  $x_0$ and $x_1$, say, and evaluates the relative probabilities of two models having predicted $x_1$ conditional on $x_0$. This typically involves evaluating $p(\theta|x_0)$ (posterior inference),  computing the integral 
\begin{equation}
    p(x_1|M_i,x_0)=\int p(x_1|\theta,x_0) p(\theta|x_0) \ {\rm d}\theta
\end{equation}
for $M_0$ and $M_1$, and then computing the posterior odds ratio 
\begin{equation}
    K_\mathrm{PPT}=\frac{p(x_1|M_1,x_0)}{p(x_1|M_0,x_0)}.
\end{equation}
The difficulty is that all the challenges of computing the evidence ratio with conventional methods, which motivated the present study, also apply to $K_\mathrm{PPT}$.

But it is straightforward estimate $K_\mathrm{PPT}$ using our methodology. The simplest way to see this is to write
\begin{equation}
    K_\mathrm{PPT}=\frac{p(x_1|M_1,x_0)}{p(x_1|M_0,x_0)}=
    \frac{p(x_0,x_1|M_1)}{p(x_0,x_1|M_0)} \frac{p(x_0|M_0)}{p(x_0|M_1)}=
    {K(x)}/{K(x_0)},
\end{equation}
which can be computed simply by training two Evidence Networks, one to compute the Bayes factor for the full data $x$ and one for the subset $x_0$.
\\
\noindent \\ \textit{In conclusion,} we have presented a novel approach for Bayesian model comparison with a wide range of applications; it overcomes the limitations of even state-of-the-art classical computational methods.

\vspace{7.cm}
\section*{Code availability}
\vspace{-0.2cm}
{A code demonstrating Evidence Networks, using the time series example, is available on GitHub:} \url{https://github.com/NiallJeffrey/EvidenceNetworksDemo} 
\github

\vspace{1.cm}
\section*{Acknowledgements}
\vspace{-0.2cm}
NJ thanks Lorne Whiteway for thoughtful discussions and comments on the manuscript. NJ is supported by STFC Consolidated Grant ST/V000780/1. This work is supported by the Simons Collaboration on “Learning the Universe".

\vspace{1.cm}
\newpage

\bibliographystyle{icml_style}
\newcommand{\newblock}{}
\bibliography{main}

\begin{thebibliography}{47}
\providecommand{\natexlab}[1]{#1}
\providecommand{\url}[1]{\texttt{#1}}
\expandafter\ifx\csname urlstyle\endcsname\relax
  \providecommand{\doi}[1]{doi: #1}\else
  \providecommand{\doi}{doi: \begingroup \urlstyle{rm}\Url}\fi

\bibitem[{Alsing} et~al.(2018){Alsing}, {Wandelt}, and {Feeney}]{delfi1}
{Alsing}, J., {Wandelt}, B., and {Feeney}, S.
\newblock {Massive optimal data compression and density estimation for
  scalable, likelihood-free inference in cosmology}.
\newblock \emph{\mnras}, 477\penalty0 (3):\penalty0 2874--2885, July 2018.
\newblock \doi{10.1093/mnras/sty819}.

\bibitem[{Alsing} et~al.(2019){Alsing}, {Charnock}, {Feeney}, and
  {Wandelt}]{delfi2}
{Alsing}, J., {Charnock}, T., {Feeney}, S., and {Wandelt}, B.
\newblock {Fast likelihood-free cosmology with neural density estimators and
  active learning}.
\newblock \emph{\mnras}, 488\penalty0 (3):\penalty0 4440--4458, September 2019.
\newblock \doi{10.1093/mnras/stz1960}.

\bibitem[Amon et~al.(2022)Amon, Gruen, and {DES Collaboration}]{amon}
Amon, A., Gruen, D., and {DES Collaboration}.
\newblock Dark energy survey year 3 results: Cosmology from cosmic shear and
  robustness to data calibration.
\newblock \emph{Physical Review D}, 105\penalty0 (2), jan 2022.
\newblock \doi{10.1103/physrevd.105.023514}.
\newblock URL \url{https://doi.org/10.1103%2Fphysrevd.105.023514}.

\bibitem[Battaglia et~al.(2018)Battaglia, Hamrick, Bapst, Sanchez-Gonzalez,
  Zambaldi, Malinowski, Tacchetti, Raposo, Santoro, Faulkner, Gulcehre, Song,
  Ballard, Gilmer, Dahl, Vaswani, Allen, Nash, Langston, Dyer, Heess, Wierstra,
  Kohli, Botvinick, Vinyals, Li, and Pascanu]{battaglia2018relational}
Battaglia, P.~W., Hamrick, J.~B., Bapst, V., Sanchez-Gonzalez, A., Zambaldi,
  V., Malinowski, M., Tacchetti, A., Raposo, D., Santoro, A., Faulkner, R.,
  Gulcehre, C., Song, F., Ballard, A., Gilmer, J., Dahl, G., Vaswani, A.,
  Allen, K., Nash, C., Langston, V., Dyer, C., Heess, N., Wierstra, D., Kohli,
  P., Botvinick, M., Vinyals, O., Li, Y., and Pascanu, R.
\newblock Relational inductive biases, deep learning, and graph networks, 2018.

\bibitem[Brehmer et~al.(2019)Brehmer, Mishra-Sharma, Hermans, Louppe, and
  Cranmer]{Brehmer_2019}
Brehmer, J., Mishra-Sharma, S., Hermans, J., Louppe, G., and Cranmer, K.
\newblock Mining for dark matter substructure: Inferring subhalo population
  properties from strong lenses with machine learning.
\newblock \emph{The Astrophysical Journal}, 886\penalty0 (1):\penalty0 49, Nov
  2019.
\newblock ISSN 1538-4357.
\newblock \doi{10.3847/1538-4357/ab4c41}.
\newblock URL \url{http://dx.doi.org/10.3847/1538-4357/ab4c41}.

\bibitem[Brehmer et~al.(2020)Brehmer, Louppe, Pavez, and Cranmer]{Brehmer5242}
Brehmer, J., Louppe, G., Pavez, J., and Cranmer, K.
\newblock Mining gold from implicit models to improve likelihood-free
  inference.
\newblock \emph{Proceedings of the National Academy of Sciences}, 117\penalty0
  (10):\penalty0 5242--5249, 2020.
\newblock ISSN 0027-8424.
\newblock \doi{10.1073/pnas.1915980117}.
\newblock URL \url{https://www.pnas.org/content/117/10/5242}.

\bibitem[Carlin \& Louis(2010)Carlin and Louis]{carlin2010bayes}
Carlin, B. and Louis, T.
\newblock \emph{Bayes and Empirical Bayes Methods for Data Analysis, Second
  Edition}.
\newblock Chapman \& Hall/CRC Texts in Statistical Science. Taylor \& Francis,
  2010.
\newblock ISBN 9781420057669.
\newblock URL \url{https://books.google.fr/books?id=484r1P5o0hQC}.

\bibitem[Cranmer et~al.(2020)Cranmer, Brehmer, and Louppe]{Cranmer201912789}
Cranmer, K., Brehmer, J., and Louppe, G.
\newblock The frontier of simulation-based inference.
\newblock \emph{Proceedings of the National Academy of Sciences}, 2020.
\newblock ISSN 0027-8424.
\newblock \doi{10.1073/pnas.1912789117}.
\newblock URL \url{https://www.pnas.org/content/early/2020/05/28/1912789117}.

\bibitem[{Crenshaw} et~al.(2021){Crenshaw}, {Connolly}, and {Kalmbach}]{pzflow}
{Crenshaw}, J.~F., {Connolly}, A., and {Kalmbach}, B.
\newblock {PZFlow: normalizing flows for cosmology, with applications to
  forward modeling galaxy photometry}.
\newblock In \emph{American Astronomical Society Meeting Abstracts}, volume~53
  of \emph{American Astronomical Society Meeting Abstracts}, pp.\  230.01, June
  2021.

\bibitem[{Doux} \& {DES Collaboration}(2022){Doux} and {DES
  Collaboration}]{doux}
{Doux}, C. and {DES Collaboration}.
\newblock Dark energy survey year 3 results: cosmological constraints from the
  analysis of cosmic shear in harmonic space.
\newblock \emph{Monthly Notices of the Royal Astronomical Society},
  515\penalty0 (2):\penalty0 1942--1972, jul 2022.
\newblock \doi{10.1093/mnras/stac1826}.
\newblock URL \url{https://doi.org/10.1093%2Fmnras%2Fstac1826}.

\bibitem[{Durkan} et~al.(2019){Durkan}, {Bekasov}, {Murray}, and
  {Papamakarios}]{neuralsplines}
{Durkan}, C., {Bekasov}, A., {Murray}, I., and {Papamakarios}, G.
\newblock {Neural Spline Flows}.
\newblock \emph{arXiv e-prints}, art. arXiv:1906.04032, June 2019.
\newblock \doi{10.48550/arXiv.1906.04032}.

\bibitem[Fenton et~al.(2016)Fenton, Neil, and Berger]{bayes_law}
Fenton, N., Neil, M., and Berger, D.
\newblock Bayes and the law.
\newblock \emph{Annual Review of Statistics and Its Application}, 3\penalty0
  (1):\penalty0 51--77, 2016.
\newblock \doi{10.1146/annurev-statistics-041715-033428}.
\newblock URL \url{https://doi.org/10.1146/annurev-statistics-041715-033428}.
\newblock PMID: 27398389.

\bibitem[Feroz(2013)]{6753897}
Feroz, F.
\newblock Calculation and applications of bayesian evidence in astrophysics and
  particle physics phenomenology.
\newblock In \emph{2013 IEEE 13th International Conference on Data Mining
  Workshops}, pp.\  8--15, 2013.
\newblock \doi{10.1109/ICDMW.2013.21}.

\bibitem[Feroz et~al.(2009)Feroz, Hobson, and Bridges]{multinest}
Feroz, F., Hobson, M.~P., and Bridges, M.
\newblock {MultiNest}: an efficient and robust bayesian inference tool for
  cosmology and particle physics.
\newblock \emph{Monthly Notices of the Royal Astronomical Society},
  398\penalty0 (4):\penalty0 1601--1614, oct 2009.
\newblock \doi{10.1111/j.1365-2966.2009.14548.x}.
\newblock URL \url{https://doi.org/10.1111%2Fj.1365-2966.2009.14548.x}.

\bibitem[Gelman et~al.(2013)Gelman, Carlin, Stern, Dunson, Vehtari, and
  Rubin]{gelman2013bayesian}
Gelman, A., Carlin, J., Stern, H., Dunson, D., Vehtari, A., and Rubin, D.
\newblock \emph{Bayesian Data Analysis, Third Edition}.
\newblock Chapman \& Hall/CRC Texts in Statistical Science. Taylor \& Francis,
  2013.
\newblock ISBN 9781439840955.

\bibitem[Han \& Carlin(2001)Han and Carlin]{doi:10.1198/016214501753208780}
Han, C. and Carlin, B.~P.
\newblock Markov chain monte carlo methods for computing bayes factors.
\newblock \emph{Journal of the American Statistical Association}, 96\penalty0
  (455):\penalty0 1122--1132, 2001.
\newblock \doi{10.1198/016214501753208780}.
\newblock URL \url{https://doi.org/10.1198/016214501753208780}.

\bibitem[Handley et~al.(2015)Handley, Hobson, and Lasenby]{polychord}
Handley, W.~J., Hobson, M.~P., and Lasenby, A.~N.
\newblock polychord: next-generation nested sampling.
\newblock \emph{Monthly Notices of the Royal Astronomical Society},
  453\penalty0 (4):\penalty0 4385--4399, sep 2015.
\newblock \doi{10.1093/mnras/stv1911}.
\newblock URL \url{https://doi.org/10.1093%2Fmnras%2Fstv1911}.

\bibitem[Handley et~al.(2019)Handley, Lasenby, Peiris, and
  Hobson]{Handley_2019}
Handley, W.~J., Lasenby, A.~N., Peiris, H.~V., and Hobson, M.~P.
\newblock Bayesian inflationary reconstructions from planck 2018 data.
\newblock \emph{Physical Review D}, 100\penalty0 (10), nov 2019.
\newblock \doi{10.1103/physrevd.100.103511}.
\newblock URL \url{https://doi.org/10.1103%2Fphysrevd.100.103511}.

\bibitem[Hochreiter \& Schmidhuber(1997)Hochreiter and Schmidhuber]{lstm}
Hochreiter, S. and Schmidhuber, J.
\newblock Long short-term memory.
\newblock \emph{Neural computation}, 9\penalty0 (8):\penalty0 1735--1780, 1997.

\bibitem[Ioffe \& Szegedy(2015)Ioffe and Szegedy]{ioffe2015batch}
Ioffe, S. and Szegedy, C.
\newblock Batch normalization: Accelerating deep network training by reducing
  internal covariate shift, 2015.

\bibitem[Jasa \& Xiang(2012)Jasa and Xiang]{acoustics}
Jasa, T. and Xiang, N.
\newblock Nested sampling applied in bayesian room-acoustics decay analysis.
\newblock \emph{The Journal of the Acoustical Society of America},
  132:\penalty0 3251--62, 11 2012.
\newblock \doi{10.1121/1.4754550}.

\bibitem[Jaynes(2003)]{jaynes07}
Jaynes, E.~T.
\newblock \emph{Probability theory: the logic of science}.
\newblock Cambridge University Press, 2003.
\newblock ISBN 0521592712.

\bibitem[{Jeffrey} \& {Wandelt}(2020){Jeffrey} and {Wandelt}]{momentnets}
{Jeffrey}, N. and {Wandelt}, B.~D.
\newblock {Solving high-dimensional parameter inference: marginal posterior
  densities \& Moment Networks}.
\newblock \emph{arXiv e-prints}, art. arXiv:2011.05991, November 2020.
\newblock \doi{10.48550/arXiv.2011.05991}.

\bibitem[Jeffrey et~al.(2020)Jeffrey, Alsing, and Lanusse]{des_lfi}
Jeffrey, N., Alsing, J., and Lanusse, F.
\newblock {Likelihood-free inference with neural compression of DES SV weak
  lensing map statistics}, Nov 2020.
\newblock ISSN 0035-8711.
\newblock URL \url{https://doi.org/10.1093/mnras/staa3594}.
\newblock staa3594, arXiv:2009.08459.

\bibitem[{Jeffrey} et~al.(2021){Jeffrey}, {Gatti}, and {DES
  Collaboration}]{mass_map}
{Jeffrey}, N., {Gatti}, M., and {DES Collaboration}.
\newblock {Dark Energy Survey Year 3 results: Curved-sky weak lensing mass map
  reconstruction}.
\newblock \emph{\mnras}, 505\penalty0 (3):\penalty0 4626--4645, August 2021.
\newblock \doi{10.1093/mnras/stab1495}.

\bibitem[Jeffreys(1998)]{jeffreys1998theory}
Jeffreys, H.
\newblock \emph{The Theory of Probability}.
\newblock Oxford Classic Texts in the Physical Sciences. OUP Oxford, 1998.
\newblock ISBN 9780191589676.

\bibitem[Johnson(2013)]{Johnson2013RevisedSF}
Johnson, V.~E.
\newblock Revised standards for statistical evidence.
\newblock \emph{Proceedings of the National Academy of Sciences}, 110:\penalty0
  19313 -- 19317, 2013.

\bibitem[Kass \& Raftery(1995)Kass and
  Raftery]{doi:10.1080/01621459.1995.10476572}
Kass, R.~E. and Raftery, A.~E.
\newblock Bayes factors.
\newblock \emph{Journal of the American Statistical Association}, 90\penalty0
  (430):\penalty0 773--795, 1995.
\newblock \doi{10.1080/01621459.1995.10476572}.
\newblock URL
  \url{https://www.tandfonline.com/doi/abs/10.1080/01621459.1995.10476572}.

\bibitem[Keysers et~al.(2020)Keysers, Gazzola, and
  Wagenmakers]{keysers2020using}
Keysers, C., Gazzola, V., and Wagenmakers, E.-J.
\newblock Using bayes factor hypothesis testing in neuroscience to establish
  evidence of absence.
\newblock \emph{Nature neuroscience}, 23\penalty0 (7):\penalty0 788--799, 2020.

\bibitem[Kingma \& Ba(2017)Kingma and Ba]{adam}
Kingma, D.~P. and Ba, J.
\newblock Adam: A method for stochastic optimization, 2017.

\bibitem[{Knuth} et~al.(2014){Knuth}, {Habeck}, {Malakar}, {Mubeen}, and
  {Placek}]{2014arXiv1411.3013K}
{Knuth}, K.~H., {Habeck}, M., {Malakar}, N.~K., {Mubeen}, A.~M., and {Placek},
  B.
\newblock {Bayesian Evidence and Model Selection}.
\newblock \emph{arXiv e-prints}, art. arXiv:1411.3013, November 2014.
\newblock \doi{10.48550/arXiv.1411.3013}.

\bibitem[{Lemos} et~al.(2020){Lemos}, {Jeffrey}, {Whiteway}, {Lahav},
  {Libeskind}, and {Hoffman}]{2020arXiv201008537L}
{Lemos}, P., {Jeffrey}, N., {Whiteway}, L., {Lahav}, O., {Libeskind}, N.~I.,
  and {Hoffman}, Y.
\newblock {The sum of the masses of the Milky Way and M31: a likelihood-free
  inference approach}.
\newblock \emph{arXiv e-prints}, art. arXiv:2010.08537, October 2020.

\bibitem[Maas et~al.(2013)Maas, Hannun, Ng, et~al.]{relu}
Maas, A.~L., Hannun, A.~Y., Ng, A.~Y., et~al.
\newblock Rectifier nonlinearities improve neural network acoustic models.
\newblock In \emph{Proc. icml}. Atlanta, Georgia, USA, 2013.

\bibitem[Masnadi-Shirazi \& Vasconcelos(2008)Masnadi-Shirazi and
  Vasconcelos]{loss_classification}
Masnadi-Shirazi, H. and Vasconcelos, N.
\newblock On the design of loss functions for classification: theory,
  robustness to outliers, and savageboost.
\newblock \emph{Advances in neural information processing systems}, 21, 2008.

\bibitem[Massimi(2020)]{massimi}
Massimi, M.
\newblock \emph{A Philosopher’s Look at the Dark Energy Survey: Reflections
  on the Use of the Bayes Factor in Cosmology}.
\newblock World Scientific, 2020.
\newblock \doi{10.1142/9781786348364_0025}.
\newblock URL
  \url{https://www.worldscientific.com/doi/abs/10.1142/9781786348364_0025}.

\bibitem[Niculescu-Mizil \& Caruana(2005)Niculescu-Mizil and
  Caruana]{calibration2}
Niculescu-Mizil, A. and Caruana, R.
\newblock Obtaining calibrated probabilities from boosting.
\newblock In \emph{Proceedings of the Twenty-First Conference on Uncertainty in
  Artificial Intelligence}, UAI'05, pp.\  413–420, Arlington, Virginia, USA,
  2005. AUAI Press.
\newblock ISBN 0974903914.

\bibitem[Papamakarios \& Murray(2016)Papamakarios and Murray]{Papamakarios_lfi}
Papamakarios, G. and Murray, I.
\newblock Fast $\varepsilon$-free inference of simulation models with bayesian
  conditional density estimation.
\newblock \emph{Advances in Neural Information Processing Systems}, pp.\
  1028--1036, 2016.

\bibitem[{Radev} et~al.(2020){Radev}, {D'Alessandro}, {Mertens}, {Voss},
  {K{\"o}the}, and {B{\"u}rkner}]{evidential}
{Radev}, S.~T., {D'Alessandro}, M., {Mertens}, U.~K., {Voss}, A., {K{\"o}the},
  U., and {B{\"u}rkner}, P.-C.
\newblock {Amortized Bayesian model comparison with evidential deep learning}.
\newblock \emph{arXiv e-prints}, art. arXiv:2004.10629, April 2020.
\newblock \doi{10.48550/arXiv.2004.10629}.

\bibitem[Ramanah et~al.(2020)Ramanah, Wojtak, Ansari, Gall, and
  Hjorth]{ramanah2020dynamical}
Ramanah, D.~K., Wojtak, R., Ansari, Z., Gall, C., and Hjorth, J.
\newblock Dynamical mass inference of galaxy clusters with neural flows, 2020.

\bibitem[Rizvi et~al.(2023)Rizvi, Pettee, and Nachman]{rizvi2023learning}
Rizvi, S., Pettee, M., and Nachman, B.
\newblock Learning likelihood ratios with neural network classifiers, 2023.

\bibitem[Secco et~al.(2022)Secco, Samuroff, , and {DES Collaboration}]{secco}
Secco, L., Samuroff, S., , and {DES Collaboration}.
\newblock Dark energy survey year 3 results: Cosmology from cosmic shear and
  robustness to modeling uncertainty.
\newblock \emph{Physical Review D}, 105\penalty0 (2), jan 2022.
\newblock \doi{10.1103/physrevd.105.023515}.
\newblock URL \url{https://doi.org/10.1103%2Fphysrevd.105.023515}.

\bibitem[{Spurio Mancini} et~al.(2022){Spurio Mancini}, {Docherty}, {Price},
  and {McEwen}]{bayes_harmonic}
{Spurio Mancini}, A., {Docherty}, M.~M., {Price}, M.~A., and {McEwen}, J.~D.
\newblock {Bayesian model comparison for simulation-based inference}.
\newblock \emph{arXiv e-prints}, art. arXiv:2207.04037, July 2022.
\newblock \doi{10.48550/arXiv.2207.04037}.

\bibitem[Taylor et~al.(2019)Taylor, Kitching, Alsing, Wandelt, Feeney, and
  McEwen]{Taylor_2019}
Taylor, P.~L., Kitching, T.~D., Alsing, J., Wandelt, B.~D., Feeney, S.~M., and
  McEwen, J.~D.
\newblock Cosmic shear: Inference from forward models.
\newblock \emph{Physical Review D}, 100\penalty0 (2), Jul 2019.
\newblock ISSN 2470-0029.
\newblock \doi{10.1103/physrevd.100.023519}.
\newblock URL \url{http://dx.doi.org/10.1103/PhysRevD.100.023519}.

\bibitem[Wakefield(2009)]{wakefield2009bayes}
Wakefield, J.
\newblock Bayes factors for genome-wide association studies: comparison with
  p-values.
\newblock \emph{Genetic Epidemiology: The Official Publication of the
  International Genetic Epidemiology Society}, 33\penalty0 (1):\penalty0
  79--86, 2009.

\bibitem[{Xu} et~al.(2015){Xu}, {Wang}, {Chen}, and {Li}]{leakyrelu}
{Xu}, B., {Wang}, N., {Chen}, T., and {Li}, M.
\newblock {Empirical Evaluation of Rectified Activations in Convolutional
  Network}.
\newblock \emph{arXiv e-prints}, art. arXiv:1505.00853, May 2015.
\newblock \doi{10.48550/arXiv.1505.00853}.

\bibitem[Yuen(2010)]{yuen2010recent}
Yuen, K.-V.
\newblock Recent developments of bayesian model class selection and
  applications in civil engineering.
\newblock \emph{Structural Safety}, 32\penalty0 (5):\penalty0 338--346, 2010.

\bibitem[Zadrozny \& Elkan(2002)Zadrozny and Elkan]{calibration1}
Zadrozny, B. and Elkan, C.
\newblock Transforming classifier scores into accurate multiclass probability
  estimates.
\newblock In \emph{Proceedings of the Eighth ACM SIGKDD International
  Conference on Knowledge Discovery and Data Mining}, KDD '02, pp.\  694–699,
  New York, NY, USA, 2002. Association for Computing Machinery.
\newblock ISBN 158113567X.
\newblock \doi{10.1145/775047.775151}.
\newblock URL \url{https://doi.org/10.1145/775047.775151}.

\end{thebibliography}

\newpage 
\appendix

\vspace{-0.3cm}

\section{Network Architecture \& recommendations} \label{append:network}
\vspace{-0.2cm}
The below table summarizes the simple default network architecture used in the demonstrations. The network is a dense network, with a few skip connections, using leaky ReLU~\citep{relu, leakyrelu} and Batch Normalization~\citep{ioffe2015batch}. The default training uses the Adam optimizer~\citep{adam} with an initial learning rate of $10^{-4}$ with an exponential decay rate of $0.95$. Unless otherwise stated, this architecture is used for the examples in the main body of the paper.

{This architecture was chosen to be simple in order to show the general applicability of Evidence Networks (i.e. they do not require architecture tuning to perform well). For the user we still recommend choosing an architecture best suited for the given data/task. In the typical manner, Network architecture can be tuned, or optimized, to give the lowest validation loss.}

{To choose a non-standard value of $\alpha$ for the l-POP loss function (i.e. $\alpha \neq 2$), the validation loss cannot be used; varying $\alpha$ effectively changes the interpretation of the loss.}

{A choice to vary $\alpha$ may be because the default choice leads to: (i) unreliable network training, with over-fitting and high-variation within an ensemble, or (ii) poor coverage test results. In case (i) it is straightforward to vary $\alpha$ (we recommend lowering $\alpha$) to improve over-fitting and consistency within the ensemble. In both cases though, the coverage test is the gold standard test. As demonstrated with DES data in section~\ref{sec:des}, a coverage test is successful if the rescaled probability residuals, $[\mathrm{model\ fraction} - p(M_1 | x)] / \sigma_{\rm binomial}$, are consistent with having mean zero and standard deviation of one.} 

{\footnotesize
\begin{verbatim}
Default (simple) architecture                              (width1 = input_shape * 1.1 + 20)
Layer (type)                    Output Shape       Connected to                     
============================================================================================
input_1 (InputLayer)            [(, 100)]   
____________________________________________________________________________________________
dense (Dense)                   (,width1)        input_1[0][0]                       
____________________________________________________________________________________________
leaky_re_lu (LeakyReLU)         (,width1)        dense[0][0]                         
____________________________________________________________________________________________
batch_normalization(BatchNorm)  (,width1)        leaky_re_lu[0][0]                   
____________________________________________________________________________________________
dense_1 (Dense)                 (, 16)           batch_normalization[0][0]           
____________________________________________________________________________________________
leaky_re_lu_1 (LeakyReLU)       (, 16)           dense_1[0][0]                       
____________________________________________________________________________________________
batch_normalization_1(BatchNorm)(, 16)           leaky_re_lu_1[0][0]                 
____________________________________________________________________________________________
dense_2 (Dense)                 (, 16)           batch_normalization_1[0][0]         
____________________________________________________________________________________________
leaky_re_lu_2 (LeakyReLU)       (, 16)           dense_2[0][0]                       
____________________________________________________________________________________________
batch_normalization_2(BatchNorm)(, 16)           leaky_re_lu_2[0][0]                 
____________________________________________________________________________________________
dense_3 (Dense)                 (, 16)           batch_normalization_2[0][0]         
____________________________________________________________________________________________
leaky_re_lu_3 (LeakyReLU)       (, 16)           dense_3[0][0]                       
____________________________________________________________________________________________
tf_op_layer_AddV2 (TensorFlowOp [(, 16)]         leaky_re_lu_3[0][0]              
                                                 batch_normalization_1[0][0]      
____________________________________________________________________________________________
batch_normalization_3(BatchNorm)(, 16)           tf_op_layer_AddV2[0][0]             
____________________________________________________________________________________________
dense_4 (Dense)                 (, 16)           batch_normalization_3[0][0]         
____________________________________________________________________________________________
leaky_re_lu_4 (LeakyReLU)       (, 16)           dense_4[0][0]                       
____________________________________________________________________________________________
dense_5 (Dense)                 (, 1)            leaky_re_lu_4[0][0]              
============================================================================================   
\end{verbatim}}

\end{document}